%% file: main.tex
\documentclass{article}

\usepackage{microtype}
\usepackage{graphicx}
\usepackage{subfigure}
\usepackage{booktabs} 

\usepackage{hyperref}

\usepackage[accepted]{icml2023}

\usepackage{amsmath}
\usepackage{amssymb}
\usepackage{mathtools}
\usepackage{amsthm}

\usepackage[capitalize,noabbrev]{cleveref}

\theoremstyle{plain}
\newtheorem{theorem}{Theorem}[section]

\newtheorem{lemma}[theorem]{Lemma}

\theoremstyle{definition}
\newtheorem{definition}[theorem]{Definition}

\theoremstyle{remark}

\usepackage[textsize=tiny]{todonotes}

\icmltitlerunning{Recovery Bounds on Class-Based Optimal Transport}

\usepackage{amsfonts,mathrsfs,bm,subfigure}
\usepackage{ifthen}

\input{math_commands.tex}

\usepackage{algorithm,algorithmic,graphicx}
\usepackage{nicefrac}
\usepackage{xcolor}
\usepackage{amsthm}
\usepackage{amssymb}

\begin{document}

\input{preamble}
\input{ashkan-macros}
\twocolumn[
\icmltitle{Recovery Bounds on Class-Based Optimal Transport: A Sum-of-Norms Regularization Framework}



\icmlsetsymbol{equal}{*}

\begin{icmlauthorlist}
\icmlauthor{Arman Rahbar}{Chalmers}
\icmlauthor{Ashkan Panahi}{Chalmers}
\icmlauthor{Morteza Haghir Chehreghani}{Chalmers}
\icmlauthor{Devdatt Dubhashi}{Chalmers}
\icmlauthor{Hamid Krim}{NCState}
\end{icmlauthorlist}

\icmlaffiliation{Chalmers}{Department of Computer Science and Engineering, Chalmers University of Technology, SE-412 96
Gothenburg, Sweden}
\icmlaffiliation{NCState}{Electrical and Computer Engineering Department, North Carolina State University, Raleigh, NC 27606, USA}

\icmlcorrespondingauthor{Arman Rahbar}{armanr@chalmers.se}

\icmlkeywords{Machine Learning, ICML}

\vskip 0.3in
]
%
%
\printAffiliationsAndNotice{} 

\begin{abstract}
  We develop a novel theoretical framework for understating OT schemes respecting a class structure. For this purpose, we propose a convex OT program with a sum-of-norms regularization term, which provably recovers the underlying class structure under geometric assumptions. Furthermore, we derive an accelerated proximal algorithm with a closed-form projection and proximal operator scheme, thereby affording a more scalable algorithm for computing optimal transport plans. We provide a novel argument for the uniqueness of the optimum even in the absence of strong convexity. Our experiments show that the new regularizer not only results in a better preservation of the class structure in the data but also yields additional robustness to the data geometry, compared to previous regularizers.
\end{abstract}

\section{Introduction}
\input{intro}
\section{Problem Formulation  and SON-regularized Optimal Transport}
\label{sec:optson}
\subsection{Problem Formulation}
\label{sec:problem}
\input{problem.tex}
\subsection{Optimal Transport with SON}
\input{OT-reg}

\section{Main Theoretical Results}
\input{theory_new.tex}

\section{Stochastic Incremental  Algorithms}
\label{sec:alg}
\input{alg3}


\section{Experiments}
\label{sec:exp}
\input{experiments}

\section{Conclusion}
 We introduced a novel theoretical framework for OT in the presence of a multi-class structure in the two domains. Accordingly, we provided first theoretical guarantees for the recovery of the class structure, and developed constrained incremental algorithms which are generally suitable for non-smooth problems  and enjoy theoretical convergence guarantees. Our experimental studies have substantiated the effectiveness of our proposed approach  in different illustrative settings and datasets. 

\section*{Acknowledgements}
This work was partially supported by the Wallenberg AI, Autonomous Systems and Software Program (WASP) funded by the Knut and Alice Wallenberg Foundation. We would like to thank the anonymous reviewers for their constructive comments.
\bibliographystyle{apalike}
\setlength{\itemindent}{-\leftmargin}
\makeatletter\renewcommand{\@biblabel}[1]{}\makeatother
\bibliography{refs-2} 
\clearpage
\appendix
\onecolumn
\input{supplement}
\input{unique}
\input{algorithm_supp.tex}

\newpage
\end{document}

%% file: math_commands.tex
\usepackage{amsmath,amsfonts,bm}
\usepackage{amssymb}
\usepackage{amsthm}

\def\eqref#1{equation~\ref{#1}}

\def\1{\bm{1}}

\DeclareMathAlphabet{\mathsfit}{\encodingdefault}{\sfdefault}{m}{sl}
\SetMathAlphabet{\mathsfit}{bold}{\encodingdefault}{\sfdefault}{bx}{n}

\def\calF{{\mathcal{F}}}

\def\calM{{\mathcal{M}}}
\def\calN{{\mathcal{N}}}

\def\calS{{\mathcal{S}}}
\def\calT{{\mathcal{T}}}

\def\calY{{\mathcal{Y}}}

\usepackage{etoolbox}

\numberwithin{equation}{section}

\newboolean{showcomments}
\setboolean{showcomments}{true}

%% file: preamble.tex
\newcommand{\RR}{{\mathbb R}}
\newcommand{\bC}{{\bf C}}
\newcommand{\bK}{{\bf K}}
\newcommand{\cP}{{\cal P}}
\newcommand{\cL}{{\cal L}}
\newcommand{\cF}{{\cal F}}
\newcommand{\cH}{{\cal H}}
\newcommand{\cXA}{{\cal X}^{(1)}}
\newcommand{\cXB}{{\cal X}^{(2)}}
\newcommand{\bxa}{{\bf x}^{(1), i}}

\newcommand{\ya}{{y}^{(1), i}}
\newcommand{\bxb}{{\bf x}^{(2), j}}


%% file: ashkan-macros.tex
\newcommand{\bGamma}{\bm{\Gamma}}
\newcommand{\bSigma}{\bm{\Sigma}}
\newcommand{\bOmega}{\bm{\Omega}}


\newcommand{\bdelta}{\bm{\delta}}
\newcommand{\bkappa}{\bm{\kappa}}
\newcommand{\bomega}{\bm{\omega}}
\newcommand{\bgamma}{\bm{\gamma}}
\newcommand{\bepsilon}{\bm{\epsilon}}
\newcommand{\blambda}{\bm{\lambda}}
\newcommand{\btheta}{\bm{\theta}}
\newcommand{\bpsi}{\bm{\psi}}
\newcommand{\bmeta}{\bm{\eta}}
\newcommand{\bzeta}{\bm{\zeta}}
\newcommand{\bmu}{\bm{\mu}}
\newcommand{\bnu}{\bm{\nu}}
\newcommand{\bpi}{\bm{\pi}}
\newcommand{\bsigma}{\bm{\sigma}}

\newcommand{\tilDelta}{\tilde{\Delta}}
\newcommand{\tlDelta}{\tilde{\Delta}}
\newcommand{\tlepsilon}{\tilde{\epsilon}}
\newcommand{\tltheta}{\tilde{\theta}}

\newcommand{\bA}{\mathbf{A}}
\newcommand{\bB}{\mathbf{B}}
\newcommand{\bD}{\mathbf{D}}
\newcommand{\bE}{\mathbf{E}}
\newcommand{\bF}{\mathbf{F}}
\newcommand{\bG}{\mathbf{G}}
\newcommand{\bH}{\mathbf{H}}
\newcommand{\bI}{\mathbf{I}}
\newcommand{\bJ}{\mathbf{J}}
\newcommand{\bL}{\mathbf{L}}
\newcommand{\bM}{\mathbf{M}}
\newcommand{\bN}{\mathbf{N}}
\newcommand{\bP}{\mathbf{P}}
\newcommand{\bQ}{\mathbf{Q}}
\newcommand{\bR}{\mathbf{R}}
\newcommand{\bS}{\mathbf{S}}
\newcommand{\bT}{\mathbf{T}}
\newcommand{\bW}{\mathbf{W}}
\newcommand{\bX}{\mathbf{X}}
\newcommand{\bY}{\mathbf{Y}}
\newcommand{\bZ}{\mathbf{Z}}

\newcommand{\bone}{\mathbf{1}}

\newcommand{\ba}{\mathbf{a}}
\newcommand{\bb}{\mathbf{b}}
\newcommand{\bc}{\mathbf{c}}
\newcommand{\bd}{\mathbf{d}}
\newcommand{\be}{\mathbf{e}}
\newcommand{\mbf}{\mathbf{f}}
\newcommand{\bg}{\mathbf{g}}
\newcommand{\bh}{\mathbf{h}}
\newcommand{\bl}{\mathbf{l}}
\newcommand{\bn}{\mathbf{n}}
\newcommand{\bp}{\mathbf{p}}
\newcommand{\bq}{\mathbf{q}}
\newcommand{\br}{\mathbf{r}}
\newcommand{\bs}{\mathbf{s}}
\newcommand{\bu}{\mathbf{u}}
\newcommand{\bv}{\mathbf{v}}
\newcommand{\bw}{\mathbf{w}}
\newcommand{\bx}{\mathbf{x}}
\newcommand{\by}{\mathbf{y}}
\newcommand{\bz}{\mathbf{z}}

\newcommand{\scrS}{\mathscr{S}}
\newcommand{\scrF}{\mathscr{F}}
\newcommand{\scrG}{\mathscr{G}}

\newcommand{\pri}{{i^\prime}}
\newcommand{\prj}{{j^\prime}}

\newcommand{\pralpha}{{\alpha^\prime}}
\newcommand{\prbeta}{{\beta^\prime}}

\newcommand{\hbeta}{\hat{\beta}}
\newcommand{\htheta}{\hat{\theta}}
\newcommand{\hsigma}{\hat{\sigma}}

\newcommand{\hp}{\hat{p}}
\newcommand{\hr}{\hat{r}}
\newcommand{\hs}{\hat{s}}
\newcommand{\hx}{\hat{x}}

\newcommand{\hN}{\hat{N}}

\newcommand{\hbSigma}{\hat{\bm{\Sigma}}}

\newcommand{\hba}{\hat{\mathbf{a}}}
\newcommand{\hbs}{\hat{\mathbf{s}}}
\newcommand{\hbx}{\hat{\mathbf{x}}}
\newcommand{\hbv}{\hat{\mathbf{v}}}

\newcommand{\hbW}{\hat{\mathbf{W}}}

\newcommand{\dif}{{\textrm d}}

\newcommand{\bbC}{\mathbb{C}}
\newcommand{\bbE}{\mathbb{E}}
\newcommand{\bbR}{\mathbb{R}}
\newcommand{\bbN}{\mathbb{N}}
\newcommand{\bbZ}{\mathbb{Z}}

\newcommand{\tlA}{\tilde{A}}
\newcommand{\tlC}{\tilde{C}}
\newcommand{\tlD}{\tilde{D}}
\newcommand{\tlF}{\tilde{F}}
\newcommand{\tlS}{\tilde{S}}

\newcommand{\tlv}{\tilde{v}}
\newcommand{\tls}{\tilde{s}}

\newcommand{\barm}{\bar{m}}
\newcommand{\barg}{\bar{g}}
\newcommand{\barh}{\bar{h}}
\newcommand{\barn}{\bar{n}}
\newcommand{\barp}{\bar{p}}
\newcommand{\barq}{\bar{q}}
\newcommand{\barr}{\bar{r}}
\newcommand{\barx}{\bar{x}}
\newcommand{\bary}{\bar{y}}

\newcommand{\barC}{\bar{C}}
\newcommand{\barH}{\bar{H}}
\newcommand{\barK}{\bar{K}}
\newcommand{\barL}{\bar{L}}

\newcommand{\barba}{\bar{\ba}}
\newcommand{\barbg}{\bar{\bg}}
\newcommand{\barbh}{\bar{\bh}}
\newcommand{\barbx}{\bar{\bx}}
\newcommand{\barby}{\bar{\by}}
\newcommand{\barbz}{\bar{\bz}}

\newcommand{\tlbA}{\tilde{\bA}}
\newcommand{\tlbD}{\tilde{\bD}}
\newcommand{\tlbE}{\tilde{\bE}}

\newcommand{\tlbW}{\tilde{\bW}}

\newcommand{\tlbv}{\tilde{\bv}}

\newcommand{\tc}{{\textrm c}}
\newcommand{\td}{{{\textrm d}}}

\newcommand{\bzero}{\mathbf{0}}

\newcommand{\suml}{\sum\limits}
\newcommand{\minl}{\min\limits}
\newcommand{\maxl}{\max\limits}
\newcommand{\infl}{\inf\limits}
\newcommand{\supl}{\sup\limits}
\newcommand{\liml}{\lim\limits}
\newcommand{\intl}{\int\limits}
\newcommand{\bigcupl}{\bigcup\limits}
\newcommand{\bigcapl}{\bigcap\limits}

\newcommand{\opconv}{{\textrm conv}}

\newcommand{\eref}[1]{(\ref{#1})}

\newcommand{\sinc}{{\textrm sinc}}
\newcommand{\tr}{{\textrm{Tr}}}
\newcommand{\var}{{\textrm Var}}
\newcommand{\cov}{{\textrm Cov}}
\newcommand{\tth}{{\textrm{th}}}
\newcommand{\st}{{\textrm st}}

\newcommand{\nwl}{\nonumber\\}

\newenvironment{vect}{\left[\begin{array}{c}}{\end{array}\right]}
\theoremstyle{definition}
\newtheorem{defi}{Definition}
\theoremstyle{remark}

%% file: intro.tex
 
Optimal transport (OT) is a classical mathematical  discipline for discovering a transport map from a source distribution to a target distribution with a minimum cost of the transport. 
It has recently been successfully used in various applications in computer vision, texture analysis, tomographic reconstruction and clustering, as documented in the recent surveys \cite{KPTSR17} and \cite{Sol18}. In many of these applications, OT exploits the geometry of the underlying spaces to effectively yield improved performance over the alternative of obviating it. 

The main purpose of this paper is to provide a theoretical foundation for OT in such geometrically-aware conditions. We focus on a scenario where a common, potentially hidden class structure is present in both domains. Examples are abundant, such as \cite{long2022video} and \cite{ott2022domain}, where respecting the class structure can significantly improve the performance. In our setup, classes are associated with well separated regions of the data space, called \emph{components}, on which the source and target distributions are supported. Each component is associated with a class and hence there is a correspondence between the components of the source and target domains. Such a model may become relevant after a suitable re-representation (embedding) of the data in a latent space. 

Our central question  is to identify conditions on the geometry of the components, under which OT can be performed  in a polynomial time, with an additional property that each source sample is mapped to its corresponding component in the target domain. For this purpose, we introduce and study a two-stage procedure, where in the first stage the components and their association is recovered and in the second one, OT is exclusively performed over the corresponding component pairs. Any off-the-shelf OT method can be used in the second stage, and hence our results mainly pertain to the recovery condition in the first stage. For this, we introduce a novel convex optimization framework, combining two popular procedures: the celebrated Kantorovich relaxation scheme of OT and the well-known sum-of-norms (SON) formulation for vector clustering \cite{LOL11,HVBJ11}, used as a regularization. Accordingly, we show for the first time that sufficiently well-separated and associated components in the two domains can be recovered by a regularized OT scheme, which naturally enjoys a polynomial algorithm due to convexity. No such results, to the best of our knowledge, are known for other regularizers. We also experimentally show that our regularizer does not only yield a better class structure preservation, but also provides additional robustness compared to other class-based regularizers

{\bf Computational benefits: 
} Despite a convex nature, the improvements of OT often come at a significant computational cost. We further argue that the SON regularization of OT also enjoys practical computational benefits by proposing a novel stochastic incremental algorithm.
First, we construct an abstract stochastic framework that is based on a combination of proximal and projection iterations, for which we give a generic proof of convergence at rate $O(1/T)$. 
Subsequently, we specialize this general scheme 
for our problem, which leads to an algorithm with  computationally low-cost iterations. Beyond the proposed SON-based framework, our proximal scheme can be used to avoid the reported convergence difficulties of gradient-based methods \cite{PN18}.

{\bf Summary of contributions}: Our main contributions can be summarized as follows:
\noindent
i. In section \ref{sec:problem}, we develop a theoretical framework for class-based OT, where we introduce the concept of multi-class recovery schemes.\\
ii. As an instance of a multi-class recovery scheme, we propose in Section~\ref{sec:regson} a new regularized formulation of OT that 
recovers a class structure typically arising in real-world problems.\\
iii. We derive the first rigorous results for recovering an OT plan that respects class structure, presented in section~\ref{theory:class_based}, with more details provided in the appendix (section \ref{sec:complete_theory}).\\
iv. We develop in Section~\ref{sec:alg} a general accelerated stochastic incremental proximal-projection optimization scheme, for which 
we give a proof of convergence at a rate $O(1/T)$ without a decaying step size. We specialize the general scheme with an explicit closed form of proximal operators and fast projections to yield a scalable stochastic incremental algorithm for computing our OT formulation.\\
v.  In section~\ref{sec:exp} and further in the appendix (section \ref{sec:additional_exp}), we investigate the algorithm on several synthetic and benchmark data sets, and demonstrate the benefits of the new regularizer.\\ 
vi. In the appendix (section~\ref{sec:uniqueness}), we develop a new proof for the uniqueness of the optimal solution of our convex formulation in spite  of its non-strong convexity, with wide applicability in other model recovery studies.\\



\subsection{Relation to Literature}

OT, first proposed by Monge as an analysis problem  \cite{monge1781memoire}, has become a classic topic in probability and statistics. A comprehensive introduction and theoretical framework can be found in   \cite{villani2008optimal,santambrogio2015optimal, peyre2019computational}. Theoretical works on extensions of OT have only recently received more attention. \cite{redko2017theoretical}, for example, provides an analysis of OT in the context of domain adaptation. In \cite{gordaliza2019obtaining}, OT is studied from the fairness point of view.  To the best of our knowledge, there is no study on the geometrically-aware extensions (regularizations) of OT, as this paper offers. 

{\bf Regularized OT:} A case for exploring new regularizers was made in \cite{CFTR17} in the context of domain adaptation applications. In \cite{blondel18a} the primal and dual formulations of OT are regularized with a strongly convex term, and the constraints are relaxed with smooth approximations. \cite{Dessein18} also propose a framework to solve discrete optimal transport problems with smooth convex regularization. Regularization is often introduced to promote sparsity. The L2 regularization  for example yields a sparse plan. It has notably been used in a doubly stochastic scheme in \cite{seguy2018large}. However, these techniques are not tailored to the underlying class structure of the data, which may not be known in advance. 
The issue of a hidden class structure is addressed in \cite{CFTR17} by the so-called Laplacian regularizer. However, no theoretical study is provided. We further illustrate the differences between the Laplacian regularizer and our framework in the experiments. It is also notable that our formulation differs from the \emph{partial domain adaption} \cite{Cao-pda} and \emph{robust domain adaption} \cite{Balaji2020RobustOT} settings developed to deal with outlier classes or samples. They do not generally take the class structure into account.

{\bf Computational aspects:} Much attention has  focused on efficient computational and numerical algorithms for OT, and a monograph focusing on this topic has appeared in \cite{peyre2019computational}. In \cite{pmlr-v108-guo20a}, an "accelerated primal-dual randomized coordinate descent (APDRCD)" algorithm is developed to solve the OT problem. An upper bound is also provided for the complexity of the algorithm and it is shown that it could be used for large-scale purposes. 
In \cite{CFTR17}, a generalized conditional gradient method is used to compute OT with the help of a couple of regularizers. Most notably Cuturi introduced an entropic regularizer and showed that its adoption with the Sinkhorn algorithm yields a fast computation of OT \cite{cuturi2013sinkhorn}; a theoretical guarantee that the Sinkhorn iteration computes the approximation in near linear time was also provided by \cite{AWR17}. Screenkhorn algorithm proposed in \cite{screenkhorn2019} performs screening to eliminate (and accelerate) the solution of the Sinkhorn algorithm. Another computational breakthrough was achieved by \cite{GCPB16} who gave a stochastic incremental algorithm to solve the entropic regularized OT problem. 
Proximal method has been used in OT but not in a stochastic scheme. Examples include \cite{pmlr-alvarez-melis18a} that uses notions of sub-modularity and also \cite{otproximalsplitting}. Unlike the above works, our algorithm is tailored to the class-based OT framework, and exploits proximal/ projection operators in a stochastic way. 

%% file: problem.tex
Consider two probability measures $\mu,\nu$ defined  on data spaces $\calY^s,\calY^t$, respectively referred to as the source and the target domains. Further, assume a non-negative function $d:\calY^s\times\calY^t\to\bbR_{\geq 0}$ evaluating the transport cost between the two domains. Classical Monge problem seeks a measurable transport map $T:\calY^s\to\calY^t$ such that the push-forward measure $T\sharp\mu$ of $\mu$ under $T$ coincides with $\nu$ and the expected transport cost  $\bbE\left[d(Y,T(Y))\right]$ is minimized, where $Y\sim\mu$.

We similarly define the class-based OT problem:
\begin{defi}
A \emph{$K$-class structure} is a pair of mixtures of $K$ probability measures:
\begin{equation}\label{eq:mixture}
\mu=\suml_{\alpha=1}^Kp_\alpha\mu_\alpha,\quad \nu=\suml_{\beta=1}^Kq_\beta\nu_\beta 
\end{equation}
 on $\calY^s$  and $\calY^t$, respectively, with a one-to-one correspondence $\pi$ between the components of the two domains (i.e. $\mu_\alpha,\nu_\beta$). We define a solution to the $K-$class Monge problem as a transport map $T:\calY^s\to\calY^t$ such that:
\begin{enumerate}
    \item For every corresponding pair $(\mu_\alpha,\nu_\beta)\in\pi$ we have $T\sharp\mu_\alpha=\nu_\beta$, i.e $T$ transports any component to its corresponding component.
    \item The expected transport cost $\bbE\left[d(Y,T(Y))\right]$ is minimized, where $Y\sim\mu$.
\end{enumerate} 
\end{defi}
Note that a solution $T$ of the multi-class Monge problem depends on the components and their association. We consider a case that these are not provided:
\begin{defi}
Given a transport cost function $d$, a \emph{Monge recovery scheme}  assigns to each pair $(\mu,\nu)$ of source and target distributions, a $K-$class structure $\left((\mu_\alpha,p_\alpha)_\alpha,(\nu_\beta,q_\beta)_\beta,\pi\right)$  for some $K\geq 0$, and its corresponding $K-$class Monge solution $T$ such that \eqref{eq:mixture} holds true. A recovery scheme is said to recover a family $\calF$ of multi-class structures if it assigns to any $(\mu,\nu)$ of $\calF$ its corresponding mixture components and their association in $\calF$. This naturally requires the components and their association to be unique in $\calF$. 
\end{defi}

The main purpose of this paper is to introduce a wide family $\calF$ of mixture models and a particular Monge recovery scheme recovering it in a polynomial time with the problem size.

\subsubsection{Disjoint Supports and Two-Stage Schemes} Our focus is on mixture models with disjoint supports. For this case, we make the following straightforward observation:
\newtheorem{prop}{Proposition}
\begin{prop}
Consider a multi-class structure $\left((\mu_\alpha,p_\alpha)_\alpha,(\nu_\beta,q_\beta)_\beta,\pi\right)$ and suppose that the supports $\mathscr{S}_\alpha$ of $\mu_\alpha$ are disjoint, and the same property holds for the supports $\mathscr{T}_\beta$ of $\nu_\beta$. Then the multi-class Monge solution comprises the restricted maps $T_\alpha:\calS_\alpha\to\calT_\beta$, for $(\alpha,\beta)\in\pi$, each being the solution of the conventional Monge problem for $(\mu_\alpha,\nu_\beta)$. 
\end{prop}

For such families of structures, the above observation suggests a two-stage recovery scheme: Given a pair of distributions $(\mu,\nu)$, first recover the components $(\mu_\alpha),(\nu_\beta)$ and their association $\pi$. Next, solve the conventional Monge problem over each associated pair. We adopt this strategy, and as a rich theory already exists for the conventional Monge problem, we mainly focus on the first stage, i.e. recovering the components and their association.

%% file: OT-reg.tex
\label{sec:regson}
A major challenge in optimal transport is that only a finite number of samples from each distribution is given. Otherwise the distributions are unknown. Consider two finite sets $\{\by_i^s\}_{i=1}^m,\{\by_j^t\}_{j=1}^n$ of points, respectively sampled from $\mu$ and $\nu$. Let  $\bD=(D_{ij}=d(\by^s_i,\by^t_j))$ be the $m\times n$  matrix of transport costs. 
We denote the $i^\tth$ row and $j^\tth$ column of $\bD$ by $\bd^i$ and $\bd_j$, respectively. We let the positive probability masses $\mu_i,\nu_j$ be respectively assigned to the data points $\by_i^s$ and $\by^t_j$.
In this discrete setup, the Monge problem is often solved with a linear programming relaxation scheme known as the Kantorovich problem: 
\begin{equation}\label{eq:simpleOT}
    \minl_{\bX\in B(\bmu,\bnu)}\langle\bD,\bX\rangle.
\end{equation}
Here, the variable matrix $\bX=(x_{i,j})$ is called the transport map and
$B(\bmu,\bnu) = \{ \bX \in R^{m \times n}, \bX \bone_{n^s} = \bmu, \bX^T \bone_{n^t} = \bnu\} $ is the set of all discrete coupling distributions between $\bmu$ and $\bnu$,
respectively denoting the vectors of elements $\mu_i,\nu_j$. Moreover, $\langle\bD,\bX\rangle=\mathrm{Tr}(\bD^T\bX)=\sum_{i,j}X_{ij}D_{ij}$ is the Euclidean inner product of two matrices. In an ideal case, one hopes that the optimal solution for $\bX$ become an assignment (permutation matrix) in which case it is seen to coincide with the solution of the Monge problem for the empirical distributions. 
\subsubsection{Multi-Class Recovery Scheme}
Now, we introduce our multi-class recovery scheme by the following convex optimization problem:
\begin{eqnarray}\label{eq:framework}
&\bX^*=\arg\minl_{\bX\in B(\bmu,\bnu)} \langle\bD,\bX\rangle \nonumber\\ &+ \lambda\left(\suml_{l,k}  R_{l,k}\|\bx_l-\bx_k\|_2 
+\suml_{l,k} S_{l,k}\|\bx^l-\bx^k\|_2\right),
\end{eqnarray}
where $\bx_l$ and $\bx^k$ denote the (transpose of the) $l^\tth$ row and  $k^\tth$ column of $\bX$, respectively. Compared to \eqref{eq:simpleOT}, a regularization term with a tuning parameter $\lambda>0$ is introduced, known as sum of norms (SON). SON is well-known for its clustering properties and hence \eqref{eq:framework} combines the class discovery properties of SON with OT. The positive kernel coefficients $S_{l,k}, R_{l,k}$ are also introduced to incorporate class prior information. 

The effect of the SON regularization in \eqref{eq:framework} is explained in \cite{LInd11, panahi2017clustering}. In short, it enforces many vanishing regularization terms (sparsity), hence yielding identical columns and identical rows in the solution. In other words, the resulting map $\bX^*$ after a suitable permutation of rows and columns is a block matrix with constant values in each block. The block structure reflects the discovered components in the source and target domains.
Under suitable conditions 
, the constraints of the Kanturovich problem will further force many blocks to be zero. If  each row and column will contain exactly one non-zero block (i.e. a block diagonal matrix under a correct order),  the solution further reflects an assignment $\pi$ between the components. In this way, \eqref{eq:framework} performs the first stage in the two-stage multi-class recovery scheme. This is made precise and proved in section~\ref{theory:class_based}.

The regularization parameter $\lambda$ sets a desired balance between the cluster structure and the underlying transport problem. When $\lambda=0$, \eqref{eq:framework} reduces to \eqref{eq:simpleOT}. For large values of $\lambda$, the SON regularization dominates the result. In a typical situation, $\lambda\to\infty$ results in all data in each domain assigned to the same cluster, hence a trivial transformation between single clusters in each domain. Smaller values of $\lambda$ lead to a larger number of identified classes in the solution. 

As mentioned, the kernel coefficients $R_{l,k},S_{l,k}$ are related to the prior knowledge of the components. For example if a perfect knowledge of the components in the source domain is available, we may set $R_{l,k} = 0$ if $\by^s_l$ and $\by^s_k$ belong to different components (classes), otherwise set $R_{l,k} = k_s(\by^s_l, \by^s_k)$ for a suitable (differentiable) kernel $k_s$. On the target side where no class information is ordinarily provided, we may set $S_{l,k} = k_t(\by^t_l, \by^t_k)$ for a suitable kernel $k_t$ of choice. 



%% file: theory_new.tex
\label{theory:class_based}
{\bf Geometric threshold:} To provide the main geometric intuitions of our analysis, we start by a simplified  case  with well-separated components.
Next, we will present a more extensive study, which is also used for proving the simplified result:
\begin{figure}[t]
    \centering
    \includegraphics[width=.3\textwidth]{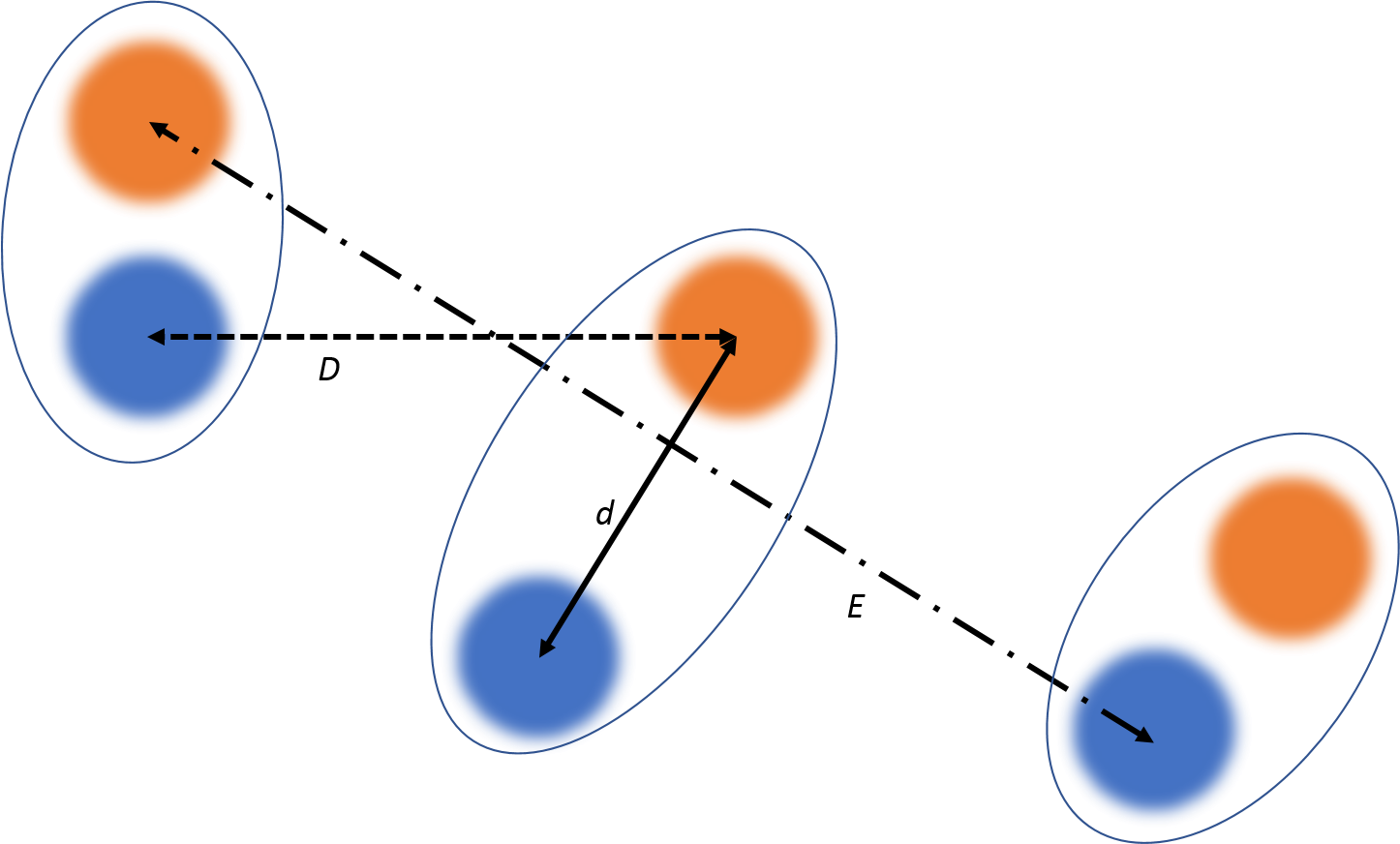}
    \caption{An example of three pairs of Gaussian clusters in the source (blue) and target (red) domains. The maximum distance $d$ between associated (paired) centers, the minimum distance $D$ between unassociated centers and the maximum distance $E$ of centers between two domains are respectively shown by solid, dashed and dash-dotted lines.}
    \label{fig:ex}
\end{figure}
\begin{theorem}\label{thm:simp}
    Consider a mixture of $K$ components in each domain with $m$ samples from each component, hence a total of $n=Km$ samples in each domain. Suppose that $(\alpha,\alpha)\in\pi$ for all $\alpha$.   
    \begin{enumerate}
        \item Consider the  $\ell_2$ distance, $d(\by_1,\by_2)=\|\by_1-\by_2\|_2$ and assume that the components are supported on spheres of radius $\omega$ and centered on $\btheta^s_\alpha,\btheta^t_\alpha$ for $\alpha=1,2,\ldots,K$, in the source and target domains, respectively. With a suitable choice of $\lambda$, the solution of \eqref{eq:framework} recovers the $K-$class structure if:
    \begin{equation}\label{eq:cond_simp2}
    \frac{D-d-2\omega}{K^{\frac 32}}\geq C\omega,
    \end{equation}
    for some universal constant $C$, where $D=\min_{\alpha\neq\beta}\|\btheta^s_\alpha-\btheta^t_\beta\|$ and $d=\max_{\alpha}\|\btheta^s_\alpha-\btheta^t_\alpha\|$.
    \item  Consider the  $\ell^2_2$ distance, $d(\by_1,\by_2)=\|\by_1-\by_2\|^2_2$. Assume that the components are  isotropic Gaussian with means $\btheta^s_\alpha,\btheta^t_\alpha$ for $\alpha=1,2,\ldots,K$, in the source and target domains, respectively. All variances are equal to $\omega^2$. With a probability higher than $1-\nicefrac{1}{n^{10}}$ the solution of \eqref{eq:framework} with a suitable choice of $\lambda$ recovers the $K-$class structure if: 
    \begin{equation}\label{eq:cond_simp}
    \frac{D^2-d^2}{K^{\frac 32}}\geq C\omega\sqrt{(E+\omega)^2+1}\log(nK),
    \end{equation}
    where $E=\max_{\alpha,\beta}\|\btheta^s_\alpha-\btheta^t_\beta\|$.
    \end{enumerate}
\end{theorem}
Fig. \ref{fig:ex} clarifies in a simple example the geometric meaning of the concepts used in the above result. As seen,  the conditions in \eqref{eq:cond_simp} and \eqref{eq:cond_simp2} require the associated clusters to be substantially closer to each other than the other clusters. 
 As expected, there are differences between the $\ell_2$ and $\ell_2^2$ geometries. For example, the scale $E$ of the problem only appears in the latter. However, in both cases we may bound the maximum number of resolvable components $K$ by a purely geometric parameter that we refer to as \emph{the geometric threshold}. In case 1), for example, defining $r=\frac{D-d-2\omega}{\omega}$, we observe that the geometric threshold is  $O(r^{\frac{2}{3}})$. Part 2) has a similar interpretation. 

{\bf Deterministic guarantee:} Now, we present an extended deterministic result that is used to prove theorem \ref{thm:simp}. In many respects, the presented results are still simplified and the most comprehensive analysis is postponed to Appendix B. 
For simplicity, 
$K$ components with equal size (number of samples) $m$ is assumed. The resulting data   partitions in the source and target domains are respectively denoted by $\{\mathscr{S}_\alpha\}$, $\{\mathscr{T}_\beta\}$. The total number of points in each domain is $n=mK$ . Further, $\mathscr{S}_\alpha$ is paired with $\{\mathscr{T}_\alpha\}$ for every $\alpha\in [K]$, i.e. $(\alpha,\alpha)\in\pi$. We investigate that the transport plan obtained by solving \eqref{eq:framework} consists of blocks, recovering both the sets of clusters $\{\mathscr{S}_\alpha\}$, $\{\mathscr{T}_\beta\}$ and their association. For this, we ensure that $X_{ij}$ remains zero for the $i^\tth$ data point in the source domain and $j^\tth$ data point in the target domain, belonging to unassociated clusters.  Accordingly, we require the \emph{ideal solution}  to be the one with $X_{i,j}=X_{\alpha,\beta}$ for $i\in \mathscr{S}_\alpha$ and $j\in \mathscr{T}_\beta$, where $X_{\alpha,\beta}$ are constants satisfying $X_{\alpha,\beta}=0$ for $\beta\neq\alpha$. 

For simplicity,  we take $S_{j,j^\prime}=1$ everywhere and study two cases where $R_{i,i^\prime}=1$ holds true either everywhere (no kernel) or for $i,i^\prime$ belonging to the same cluster and  $R_{i,i^\prime}=0$ otherwise  (perfect kernels in the source domain). The general case is presented in Appendix B. Introducing an indicator variable $R$, the first case is referred to by $R=0$ and the second one by $R=1$. Note also that we assume the ideal solution to be feasible for the optimization problem in \eqref{eq:framework}, which requires for every $i,i^\prime\in \mathscr{S}_\alpha$ and $j,j^\prime\in \mathscr{T}_\alpha$ that $\mu_i=\mu_{i^\prime}=\nu_j=\nu_{j^\prime}$. In Appendix B, we treat the general infeasible cases by considering a relaxation of \eqref{eq:framework}.

In the context of recovery  by the Kantorovich relaxation, a key concept is cyclical monotonicity \cite{villani2008optimal}, which we slightly modify and state below:
\begin{defi}	
We say that a set of coefficients $D_{\alpha,\alpha^\prime}$ for $\alpha,\alpha^\prime\in [K]$ satisfies the $\delta-$strong cyclical monotonicity condition if for each 
simple loop $\alpha_1\to \alpha_2\to \ldots \to \alpha_k\to \alpha_{k+1}=\alpha_1$ with length $k>1$ we have 
\begin{equation}\label{eq:strong_pos_loop_cond}
\suml_{l=1}^kD_{\alpha_l\alpha_{l+1}}> \suml_{l=1}^kD_{\alpha_l\alpha_{l}}+k\delta.
\end{equation}  
\end{defi}
Compared to the standard notion of cyclic monotonicity, we introduce a constant $\delta\geq 0$ on the right hand side of \eqref{eq:strong_pos_loop_cond}, which can be nonzero only when $(D_{\alpha,\beta})$ has a discrete or discontinuous nature. We apply this condition to the average distance of clusters given by 
$
D_{\alpha,\beta}= \frac{1}{m^2} \suml_{i\in \mathscr{S}_\alpha, j\in \mathscr{T}_\beta}D_{i,j}.
$
We denote by $\Delta$ the maximum of the values $\nicefrac {\left\|\bd_i-\bd_{i^\prime}\right\|}{\sqrt{n}}$ and $\nicefrac {\left\|\bd^j-\bd^{j^\prime}\right\|}{\sqrt{n}}$ where source points $i,i^\prime$ and target points $j,j^\prime$ belong to the same cluster and we remind that $\bd_i,\bd^j$ respectively refer to the rows and columns of $\bD$.
We also  define
$
\omega_\alpha :=\suml_{i\in \mathscr{S}_\alpha}\mu_i=\suml_{j\in D_{\alpha}}\nu_j
$
and then take $T_{\alpha,\beta}=\suml_{\gamma\in [K]}
\left(\frac{R\omega_\alpha}{\sqrt{\omega_\alpha^2+\omega_\gamma^2}}+
\frac{\omega_\beta}{\sqrt{\omega_\beta^2+\omega_\gamma^2}}
\right)-\frac{1+R}{\sqrt 2}$
. Finally, we define
\[
\Lambda_{\alpha,\beta}= \left(
T_{\alpha,\beta}+
\frac{\omega_\alpha+R\omega_\beta}{\sqrt{\omega_\beta^2+\omega_\beta^2}} \right)^{-1},
\]
and take $\Lambda$ as its maximum over $\alpha\neq\beta$. Accordingly, we obtain the following result:
\begin{theorem}
\label{thm:guarantee}
Suppose that
$(D_{\alpha,\beta})$ is $\delta-$strongly cyclical monotone. Take $\lambda$ such that $\Delta \leq \lambda\sqrt{\nicefrac{m}{K}}$. Then, the solution of \eqref{eq:framework} is given by $X_{ij}=X_{\alpha,\beta}$ for $i\in \mathscr{S}_\alpha$ and $j\in \mathscr{T}_\beta$ satisfying one of the following two conditions:
\begin{enumerate}
\item We have $X_{\alpha,\beta}=\nicefrac{\omega_\alpha}{m^2}\delta_{\beta,\alpha}$ if 
$
\Delta\sqrt{K}\leq\lambda\sqrt{m}
\leq \Lambda\delta
$
\item Otherwise, we have
$
\delta\suml_{\beta\neq \alpha}X_{\alpha,\beta}\leq
\lambda(1+R)\sqrt m\suml_{\alpha\neq\alpha^\prime}
\sqrt{\omega^2_\alpha+\omega^2_{\alpha^\prime}
}.
$
\end{enumerate}
Furthermore, the solution is unique in part 1 if all inequalities are strict.
\begin{proof}
Proof can be found in section \ref{proof:guarantee}.
\end{proof}
\end{theorem}
The first part of theorem \ref{thm:guarantee} establishes ideal recovery. It can be understood in light of theorem \ref{thm:simp}, e.g., in part 1. The term $\Delta$ corresponds to $\omega$ in \ref{eq:cond_simp2}. It is the maximal cluster diameter in a geometry embedded by the vectors $\bd_i,\bd^j$\footnote{Note that here $\bd_i,\bd^j$ are treated as a special embedding of the source and target points, respectively, which may be called the \emph{distance embedding}. As seen, we do not assume any inherent geometry in the two domains and instead rely on the induced $\ell_2$ geometry of the distance embedding.}. On the other hand, $\delta$ corresponds to $D-d-2\omega$ which is the gain of the assignment. The second part gives an upper bound on the error $\suml_{\beta\neq \alpha}X_{\alpha,\beta}$. Note that $\Delta$ is always smaller with $R=1$ compared to $R=0$, making the conditions less restrictive. This reflects the intuitive fact that introducing kernels simplifies the recovery.

%% file: alg3.tex
\subsection{Accelerated Proximal-Projection Scheme}
\label{sec:alg_scheme}
An important advantage of the framework in \eqref{eq:framework} is the possibility of applying stochastic optimization techniques. Since the objective term includes a large number of non-smooth SON terms, our stochastic optimization avoids calculating the (sub)gradient or the proximal operator of the entire objective function, which is numerically infeasible for large-scale problems.   
Our algorithm is obtained by introducing the following "template function":
\begin{equation}
\label{eq:phif}
\phi_{\rho,\bzeta,\bmeta}(\bp,\bq)=\langle\bp,\bzeta\rangle+\langle\bq,\bmeta\rangle
+\rho\|\bp-\bq\|_2,
\end{equation}
and noting that the objective function in \eqref{eq:framework} can be written as

\begin{equation}
\begin{aligned}
\label{eq:fullobj}
&\suml_{l\neq k}\phi_{R_{l,k},\frac 1{2(n-1)}\bd_l,\frac 1{2(n-1)}\bd_k}(\bx_l,\bx_k)+
\\
&\suml_{l\neq k}\phi_{S_{l,k},\frac 1{2(m-1)}\bd^l,\frac 1{2(m-1)}\bd^k}(\bx^l,\bx^k),
\end{aligned}
\end{equation}
with a total number of $P=m(m-1)+n(n-1)$ summands in the form of the template function. This places the problem in the setting of \emph{finite sum} optimization problems \cite{bcn18}. However, there are two obstacles to the application of stochastic optimization techniques: First,  the terms in (\ref{eq:fullobj}) are not smooth, so gradient methods do not apply and second, \eqref{eq:framework} involves a fairly complex constraint. We address these issues in the following.

{\bf Non--smooth terms:} We exploit the highly effective  proximal methodology for optimizing non--smooth functions \cite{PB14,CP11} using a proximal operator. Defazio further gives a stochastic acceleration technique using proximal operators for unconstrained problems \cite{Def16}. In addition to its fast convergence, the main advantage of this scheme is its potential constant step size  convergence in contrast  to the ordinary stochastic gradient approach. It unfortunately does not address constrained optimization problems.

{\bf Constrained optimization}: Facing a constrained optimization problem, the calculation of  the proximal operators over the feasible set is numerically intractable. However, we observe an appealing structure in the constraint which lends itself to a more efficient stochastic implementation: Recalling the definition of an $n-$dimensional standard simplex
\[
S^{(n)}=\left\{\bx=(x_i\geq 0)_{i=1}^n\mid\suml_i x_i=1\right\},
\]
we define the weighted cylinder-simplices  $S_l(\mu)=\{\bX\mid\bx_l\in\mu S^{(n)}\}$ and $S^k(\nu)=\{\bX\mid\bx^k\in\nu S^{(m)}\}$ respectively corresponding to the $l^\tth$ row and $k^\tth$ column of $\bX$ with weights $\mu,\nu\geq 0$. We then observe that the constraint set $B(\bmu,\bnu)$ is equal to $B(\bmu,\bnu)=(\bigcap_{l=1}^mS_l(\mu_l))\cap(\bigcap_{k=n}^mS^k(\nu_k))$, which is an intersection of $Q=m+n$ weighted cylinder-simplices.

In summary, the optimization problem in \eqref{eq:framework} can be written in the following abstract form:
\begin{equation}\label{eq:decompose}
\minl_{x\in\bbR^D}\suml_{p=1}^P\phi_p(x)\quad\st\quad x\in\bigcapl_{q=1}^QS_q,
\end{equation}
where each term $\phi_p$ denotes a template function term in the objective and each set $S_q$ is a weighted cylinder-simplex. The values of $P,Q$ in \eqref{eq:framework} are given above and $D=mn$.
 \cite{Bert11}, \cite{WB16} and \cite{PN18} give general stochastic incremental schemes that combine gradient, proximal and projected schemes for optimizing such finite sum problems with convex constraints.  These do not, however, use acceleration and their respective convergence is only guaranteed  with a variable and vanishing step size, which is practically difficult to control and often yields extremely slow convergence. 

{\bf Our proposed method:} We herein jointly exploit the two ideas in \cite{Def16} and \cite{WB16} to obtain an accelerated proximal scheme for constrained framework in \eqref{eq:framework}. 
Further, we shortly show in Lemma \ref{lem:proxop} that the proximal operator can be computed in closed form for our problem. Together with the projection to the simplex from \cite{condat16,DSSC08}, this gives a stochastic incremental algorithm with much less costly iterations. 

We extend the acceleration techniques of unconstrained optimization as in the Defazio's scheme (known as Point--SAGA) to the constrained setting. Point--SAGA utilizes individual "memory" vectors for each term in the objective function, which store a calculated subgradient of a selected term in every iteration. These vectors are subsequently used as an estimate of the subgradient at subsequent iterations. We extend this scheme by introducing similar memory vectors to constraints. Each memory vector $\bh_m$ for a constraint $S_m$ stores the last observed normal (separating) vector to $S_m$. At each iteration either an objective term $\phi_p$ or a constraint component $S_q$ is considered by random selection. Accordingly, we propose the following rule for updating the solution:
\begin{equation}\label{eq:alg1}
\bx_{t+1}=\left\{\begin{array}{lc}
\mathrm{prox}_{\mu\phi_{p_t}}\left(\bx_t+\mu\bg_{p_t}\right),     & \phi_{p_t}\ \text{is selected}  \\
\mathrm{proj}_{S_{q_t}}\left(\bx_t+\mu\bh_{q_t}\right)     & S_{q_t}\ \text{is selected}
\end{array}\right.,
\end{equation}
where $t$ is the iteration number, $\mu>0$ is the fixed step size and $p_t,q_t$ denote the selected index in this iteration (only one of them exists). At each iteration, the corresponding memory vector to the selected term is also updated.
Depending on the choice of $\phi_{p_t}$ or $S_{q_t}$, either $\bg_{p_t}\gets\bg_{p_t}+\ba_t$ or $\bh_{q_t}\gets\bh_{q_t}+\ba_t$, where \begin{equation}\label{eq:alg2}
\ba_t=\rho\frac{\bx_t-\bx_{t+1}}{\mu}-\alpha\left(\suml_n\bg_n+\suml_m\bh_m\right),
\end{equation}
where $\rho\in(0\ 1)$ and $\alpha>0$ are design constants. The vector $\ba_t$ consists of two parts: the first part $\rho\frac{\bx_t-\bx_{t+1}}\mu$ calculates a sub-gradient or a normal vector at point $\bx_{t+1}$ corresponding to the selected term. The second term, the sum of the memory terms, implements acceleration. Our algorithm bears marked differences with Point-SAGA. While acceleration by the sum of memory vectors is also employed in Point-SAGA, it is moved in our scheme from the update rule of $\bx_t$ to the update rule of $\bg_t$. Also, the design parameters $\rho$ and $\alpha$ are introduced to improve convergence. Similar to Point-SAGA we only need to calculate the sum of memory terms once in the beginning and later update it by simple manipulations. As we later employ initialization of the memory vectors by zero, the first summation trivially leads to zero. 
\paragraph{Convergence analysis:}
\label{sec:convergence_analysis}
\newtheorem{assum}{Assumption}
We show that for a generic convex optimization problem of the form in \eqref{eq:decompose}, the algorithmic scheme in section \ref{sec:alg_scheme} converges with a guaranteed rate, under the following mild assumptions: 
\begin{assum}
The functions $\phi_p$ are $\beta-$Lipschitz.
\end{assum}

\begin{assum}
We require the monotone inclusion problem
\begin{equation}\label{eq:first_order}
    \bzero\in\suml_{p=1}^P\partial\phi_p(\bx)+\suml_{q=1}^Q\partial I_{S_q}(\bx)
\end{equation}
to have a solution at $\bx=\bx^*$ with a finite optimal value $\phi^*$ and $\bg_p^*\in\partial\phi_p(\bx^*)$ and $\bh_q^*\in\partial I_{S_q}(\bx^*)$ satisfying $\sum_p\bg_p^*+\sum_q\bh_q^*=\bzero$. Furthermore, we assume that \begin{equation}\label{eq:assu_1}
    \sum_p\left\|\bg_p^*\right\|^2+\sum_q\left\|\bh_q^*\right\|^2=O(\beta^2R)
\end{equation}
where $R=P+Q$ is the total number of terms.
\end{assum}
 Here, $\partial\phi (\bx)$ and $\partial I_{S}(\bx)$ respectively denote the subdifferential of the function $\phi$ and the cone of normal vectors to the set $S$ at $\bx$. It is well-known that any solution to \eqref{eq:first_order} is an optimal feasible solution to \eqref{eq:decompose}.  
 \begin{assum}
 We assume that the algorithm is initialized with $\bg_p=\bh_q=\bzero$.
 \end{assum}

 Then, we can show the following result.
\begin{theorem}
\label{thm:convergence}
 Suppose that Assumption 1-3 are satisfied. Then for  $\alpha,\rho,\mu>0$ and $\alpha<2(1-\rho)$ the following holds true:
 \begin{enumerate}
     \item Defining $\barbx_t=\frac 1 t\suml_{\tau=0}^{t-1}\bx_{\tau}$ and $\eta=(1+\nicefrac QP)(\nicefrac{\|\bx_0-\bx^*\|^2}{\beta\mu}+\beta\mu R)$ we have 
     \begin{eqnarray}
         &\bbE\left[\suml_p\phi_p(\barbx_t)\right]-\phi^*\leq c\beta P\left(\frac\eta t+\sqrt{\frac{\beta\mu\eta} t}\right),\nwl &\bbE\left[\suml_q\mathrm{dist}^2(\barbx_t,S_q)\right]\leq c\beta\mu P\frac\eta t,
     \end{eqnarray}
     where $c$ is a constant depending on $\rho,\alpha$ and the underlying constant in \eqref{eq:assu_1}.
     \item Moreover,
     $   \suml_{\tau=0}^\infty\bbE[\|\bx_{\tau+1}-\bx_\tau\|^2]\leq c\frac{\mu\beta P}{P+Q}\eta.
     $
 \end{enumerate}
 \begin{proof}
 The proof is given in section \ref{proof:convergence}.
 \end{proof}
\end{theorem}

{\bf Comment}: As expected, the results only depend on $\beta\mu$, except for the optimality gap being linearly proportional to $\beta$. Applying this technique to our problem of interest and assuming that $m,n$ are of the same order, we observe that $P=O(n^2)$ and $Q=O(n)$. Since many terms in our objective function are for regularization, it is fair to consider the relative optimality gap obtained by driving the optimality gap to the number of objective terms. We observe that this quantity is controlled by  $\eta/t$. We conclude that $t\sim\eta$ iterations is required to achieve a desired relative optimality gap. The total feasibilty gap is controlled by $\beta\mu P\eta$. If we take $\beta\mu\sim\nicefrac 1n$, we obtain $\eta\sim n$ and the relative optimality gap vanishes in $O(n)$ iterations. Then, the total optimality gap and feasibility gap will vanish in $O(n^3)$ and $O(n^2)$ iterations, respectively. In the absence of the regularization terms, we may reorganize the objective to have only $O(n)$ terms. In this case, taking $\beta\mu\sim \nicefrac 1{\sqrt{n}}$, we get $\eta\sim O(\sqrt n)$ and we require $O(n^{\frac 32})$ and $O(n)$ iterations to control the total optimality and feasibility gaps. Compared to the results of \cite{guo20}, which establishes convergence in $O(n^{\frac 52})$ our convergence rates are better. Moreover, the $O(n^2)$ dependence can be improved by reducing the number of terms in the objective function. It has been pointed out that in the sum of norms approach many terms may be redundant and only  $O(n)$ terms corresponding to pre-selected pairs $(i,j)$ can be sufficient. 

\subsection{Proximal Operator for the SON-Regularized Kantorovich Relaxation}
\label{sec:proxop}
We next show that we can explicitly compute the proximal operator for each term in (\ref{eq:fullobj}):
\begin{theorem}
\label{lem:proxop}
The proximal operator of the template function $\phi_{\rho,\bzeta,\bmeta}$ is given by
$\calT_{\mu \rho}(\bp-\mu\bzeta,\bq-\mu\bmeta),
$
where
\begin{eqnarray}\label{eq:Tau}
&\calT_{\lambda}(\ba,\bb)=\left(\frac{\ba+\bb}2+\calT_{\lambda}\left(\frac{\ba-\bb}2\right),
\frac{\ba+\bb}2-\calT_{\lambda}\left(\frac{\ba-\bb}2\right)\right),
\end{eqnarray}
and $\calT_{\lambda}(\bc)$ is a thresholding operator given by $\frac{\|\bc\|-\lambda}{\|\bc\|}\bc$ if $\|\bc\|\geq\lambda$ and is zero otherwise.
\begin{proof}
Proof is found in section \ref{sec:proof_proxop}.
%
\end{proof}
\end{theorem}

%% file: experiments.tex
We now investigate the effectiveness of our OT framework for the well-known domain adaptation problem which aims at improving the performance of a model in a target domain by utilizing the knowledge available in a different but related source domain. In the appendix (section \ref{sec:additional_exp}), we investigate the benefits of several other aspects of our framework, such as early stopping, class diversity, and unsupervised domain adaptation. 
We compare our method (OT-SON) with the other regularized optimal transport-based methods OT-l1l2, OT-lpl1 and OT-Sinkhorn, as developed and used in \cite{CFTR17,cuturi2013sinkhorn,perrot2016mapping}.




\subsection{Domain Adaptation with Real-World Datasets}
\subsubsection{MNIST and USPS}
\label{sec:mnistusps}
In these experiments,  we compare the different models on the real-world images of digits. For this, we consider the MNIST data as  the source and the USPS data as the target. To further increase the difficulty of the problem,  we use all $10$ classes of the source (MNIST) data, and  we discard some of the classes of the target (USPS) data.
In our experiments, each object (image) is represented by $256$ features.
By discarding the different subsets from the USPS data, we consider several pairs of source and target datasets. i) real1: the  USPS classes are $1,2,3,5,6,7,8$, ii) real2: the USPS classes are $0,2,4,5,6,7,9$, iii) real3: the USPS classes are $0,1,3,5,7,9$, and iv) real4: the USPS classes are: $0,1,3,4,6,8,9$. 
We note that these settings where the  number of classes is different between the domains are the typical cases in practice. Therefore, our class-specific OT approach is more suitable and robust to class imbalance, as it avoids splitting a class in one domain among multiple classes in another domain.

The transformed source samples are used to train a 1-nearest neighbor classifier. We then use this (parameter-free) classifier to estimate the class labels in the target data and then compute the respective accuracy. Table \ref{tab:real_data_acc} shows the best accuracy results for different OT-based models when using different values of the regularization parameters $\lambda_1$ and $\lambda_2$ (i.e., $\lambda_1, \lambda_2 \in \{10^{-5},..., 10^3\}$)\footnote{For OT-l1l2 and OT-lpl1, $\lambda_1$ is the entropic regularization parameter and $\lambda_2$ is class regularization parameter.}. Specifically, for each OT method we use different values of regularization parameters and we report the best accuracy achieved by that method.  
We observe, i) OT-SON yields the highest accuracy scores, and ii) it is significantly more robust to variation of the regularization parameters, in comparison to the other methods. Moreover, the other methods are prone to yielding numerical errors for small regularizations.

\begin{table}[th!]
\centering
\begin{tabular} {l|l l l l}
model & real1 & real2 & real3 & real4 \\ 
\hline
OT-SON & \textbf{0.550} & \textbf{0.564} & \textbf{0.608} & \textbf{0.628}\\
OT-l1l2 & 0.421 & 0.507 & 0.500 & 0.621\\
OT-lpl1 & 0.457 & 0.521 & 0.516 & 0.592\\
OT-Sinkhorn & 0.414 & 0.521 & 0.508 & 0.621\\
\end{tabular}
 \caption{Accuracy scores 
on MNIST and USPS.
 }
\label{tab:real_data_acc}
\end{table}
\subsubsection{Caltech Office} A commonly used dataset for domain adaptation is an object recognition dataset known as \emph{Caltech Office} \cite{office_data, caltech_data} which consists of four different domains: A (Amazon, 958 samples) , W (Webcam, 295 samples), C (Caltech, 1123 samples), and D (DSLR, 157 samples). These domains have 10 classes of objects in common that are represented with two sets of features: SURF (800 features) and DeCAF (4096 features). Similar to the setting in \cite{CFTR17}, we use all possible pairs of the four domains as source and target with SURF features, and we remove the classes 3, 5, and 7 from the target domain. Similar to the previous study, we assume imbalanced source and target classes in order to make the task more realistic. 
Table \ref{tab:caltech_office} compares the classification accuracies achieved by our method with the scores obtained by the three other methods. We calculate the accuracy similar to Section \ref{sec:mnistusps}. In these experiments we use class information in the source domain together with Gaussian kernels. Specifically, we set $R_{l,k}=0$ if $y_l^s$ and $y_k^s$ are not in the same classes and otherwise we set $R_{l,k}= \lambda \exp (-\|y_l^s - y_k^s\|^2)$ for different values of $\lambda$. We observe that our method yields the best results in seven cases. In other cases, our method is still competitive compared to the alternatives. In addition, we conclude that the different methods may perform differently on different datasets. Even though the setting of imbalanced classes is more important in practice and we have focused more on that, for the sake of completeness, we also compare our method with the three other methods in a setting where the classes are balanced. We use the same number of classes in the source ($W$) and target ($C$) domains. The accuracy results for the different methods are respectively: i) OT-SON: 0.260, ii) OT-l1l2: 0.244, iii) OT-lpl1: 0.247, and iv) OT-Sinkhorn: 0.243. We observe that even in this setting, our method yields the highest accuracy.

\begin{table}[th!]
\centering
\begin{tabular} {l|l l l l}
S$\rightarrow$T & OT-SON & OT-l1l2 & OT-lpl1 & OT-Sinkhorn\\ 
\hline

A$\rightarrow$C & 0.3552 & 0.3229 & \textbf{0.3565} & 0.3018\\
A$\rightarrow$D & 0.3274 & 0.3097 & \textbf{0.3539} & 0.3008 \\
A$\rightarrow$W & \textbf{0.3144} & 0.2577 & 0.3092 & 0.2422 \\
C$\rightarrow$A & \textbf{0.4601} & 0.3714 & 0.4120 & 0.3548 \\
C$\rightarrow$D & 0.3982 & 0.3274 & \textbf{0.4424} & 0.3362\\
C$\rightarrow$W & \textbf{0.3556} & 0.2474 & 0.3144 & 0.2474 \\
D$\rightarrow$A & \textbf{0.3248} & 0.2812 & \textbf{0.3248} & 0.2812 \\
D$\rightarrow$C & 0.2720 & 0.2658 & \textbf{0.2956} & 0.2621 \\
D$\rightarrow$W & 0.7216 & \textbf{0.7628} & 0.5876 & 0.7525 \\
W$\rightarrow$A & \textbf{0.2661} & 0.2225 & 0.2616 & 0.2210 \\
W$\rightarrow$C & \textbf{0.2149} & 0.1962 & 0.2124 & 0.2012\\
W$\rightarrow$D & \textbf{0.8230} & 0.7964 & 0.6637 & 0.8141 \\

\end{tabular}
 \caption{Accuracy scores 
on Caltech Office (S: source, T: target).
 }
\label{tab:caltech_office}
\end{table}



%% file: supplement.tex
\section{Extension of Theorem \ref{thm:guarantee}}
\label{sec:complete_theory}
We consider the analysis of our proposed method for general kernel coefficient and cluster sizes. Hence, we respectively consider two partitions $\{C_\alpha\}$, $\{D_\beta\}$ of $[n], [m]$ with the same number of parts $K$. We denote the cardinalities of $C_\alpha$ and $D_\beta$ by $n_\alpha$ and $m_\beta$, respectively. Further, we consider a permutation $\pi$ on $[K]$ as the target of OT. Also, we address infeasibility by consider the following optimization:
\begin{eqnarray}
\label{eq:main_kernel}
&\minl_{\bX\in\bbR_{\geq 0}^{n\times n}}\langle\bD,\bX\rangle+\nwl
&\lambda\left(\suml_{i,i^\prime}R_{i,i^\prime}\|\bx_i-\bx_{i^\prime}\|_2+
\suml_{j,j^\prime}S_{j,j^\prime}\|\bx^j-\bx^{j^\prime}\|_2\right)\nwl
&+\frac\theta 2\left(\left\|\bX\bone-\bmu\right\|_2^2+\left\|\bX^T\bone-\bnu\right\|_2^2\right)
\end{eqnarray}
where $\theta>0$ is a design parameter and we remind that $\bx_i=(X_{i,j})_j$, $\bx^j=(X_{i,j})_i$, and $R_{i,i^\prime}$ and $S_{j,j^\prime}$ are positive kernel coefficients. Now, we introduce few intermediate optimizations to carry out the analysis. Define the following more general characteristic optimization:

\begin{eqnarray}\label{eq:char_kernel}
&\minl_{X_{\alpha,\beta}\geq 0}\suml_{\alpha,\beta}n_\alpha m_\beta X_{\alpha,\beta}D_{\alpha,\beta}+\nwl
&\lambda\left(\suml_{\alpha,\alpha^\prime}
R_{\alpha, \alpha^\prime}\|\bx_\alpha-\bx_{\alpha^\prime}\|_M+
\suml_{\beta,\beta^\prime} S_{\beta,\beta^\prime}\|\bx^\beta-\bx^{\beta^\prime}\|_N\right)\nwl
&+\frac\theta 2\left(\suml_\alpha n_\alpha\left(\ba_M^T\bx_\alpha-\mu_\alpha\right)^2
+\suml_\beta m_\beta\left(\ba_N^T\bx^\beta-\nu^\beta\right)^2\right)\nwl
&\ 
\end{eqnarray}

where
\[
R_{\alpha,\alpha^\prime}=\suml_{i\in C_\alpha,\ i^\prime\in C_{\alpha^\prime}}
R_{i,i^\prime},\quad
S_{\beta,\beta^\prime}=\suml_{j\in D_\beta,\ j^\prime\in D_{\beta^\prime}}
S_{j,j^\prime}
\]

\[
D_{\alpha,\beta}=\frac{\suml_{i\in C_\alpha, j\in D_\beta}D_{i,j}}{n_\alpha m_\beta},\ \mu_\alpha=\frac{\suml_{i\in C_\alpha}\mu_i}{n_\alpha},\ \nu^\beta=\frac{\suml_{j\in D_\beta}\nu_j}{m_\beta},
\]

$\|\bx\|_O=\sqrt{\bx^TO\bx}$, $N, M$ are diagonal matrices with $n_\alpha$ $m_\beta$ as diagonals, resepectively,  $\bx_\alpha=(X_{\alpha,\beta})_\beta$ and $\bx^\beta=(X_{\alpha,\beta})_\alpha$, and $\ba_M=(m_\alpha)_\alpha$, $\ba_N=(n_\alpha)_\alpha$.

Further, define the ideal optimization:

\begin{eqnarray}\label{eq:ideal_kernel}
&\minl_{Y_{\alpha,\beta}\geq 0}\suml_{\alpha,\beta}Y_{\alpha,\beta}D_{\alpha,\beta}
\nwl
&\mathrm{s.t}
\nwl
&q_\beta:\ \bone^T\by^\beta=\sigma^\beta, p_\alpha: \bone^T\by_\alpha=\sigma_\alpha
\end{eqnarray}

where $\sigma_\alpha=(n_\alpha\mu_\alpha+m_{\pi(\alpha)}\nu^{\pi(\alpha)})/2$, $\sigma^\beta=\sigma_{\pi^{-1}(\beta)}=(n_{\pi^{-1}(\beta)}\mu_{\pi^{-1}(\beta)}+m_\beta\nu^{\beta})/2$, and $\{p_\alpha\},\{q_\beta\}$ are dual variables. Also, define $\delta_\alpha=(\mu_\alpha n_\alpha-m_{\pi(\alpha)}\nu^{\pi(\alpha)})/2$, $\delta^\beta=-\delta_{\pi^{-1}(\beta)}=(m_\beta\nu^\beta-n_{\pi^{-1}(\beta)}\mu_{\pi^{-1}(\beta)})/2$ and $\bdelta=(\delta_\alpha)$. Finally, take
\[
R_{i,\alpha}=\sum_{i^\prime\in C_{\alpha}}R_{i,i^\prime},\quad
S_{j,\beta}=\sum_{j^\prime\in D_{\beta}}S_{j,j^\prime}
\]
In case $\bdelta=0$, we may also define the tight characteristic optimization:
\begin{eqnarray}\label{eq:char_kernel_tight}
&\minl_{X_{\alpha,\beta}\geq 0}\suml_{\alpha,\beta}n_\alpha m_\beta X_{\alpha,\beta}D_{\alpha,\beta}+\nwl
&\lambda\left(\suml_{\alpha,\alpha^\prime}
R_{\alpha, \alpha^\prime}\|\bx_\alpha-\bx_{\alpha^\prime}\|_M+
\suml_{\beta,\beta^\prime} S_{\beta,\beta^\prime}\|\bx^\beta-\bx^{\beta^\prime}\|_N\right)\nwl
&\text{s.t.}\nwl
&\ba_M^T\bx_\alpha=\mu_\alpha,
\quad\ba_N^T\bx^\beta=\nu^\beta\nwl
&\ 
\end{eqnarray}
which in this case, coincides with \eqref{eq:char_kernel} when $\theta=\infty$. We also define
\begin{eqnarray}
 &T_\alpha=\frac{1}{n_\alpha m_{\pi(\alpha)}}\suml_{\alpha^\prime\neq\alpha} \frac{R_{\alpha,\alpha^\prime}\frac{\sigma_\alpha}{n_\alpha m_{\pi(\alpha)}}}
{\sqrt{\left(\frac{\sigma_\alpha}{n_\alpha m_{\pi(\alpha)}}\right)^2
+
\left(\frac{\sigma_{\alpha^\prime}}{n_{\alpha^\prime} m_{\pi(\alpha^\prime)}}\right)^2}}
,\nwl
\nwl
&U^\beta=\frac{1}{n_{\pi^{-1}(\beta)} m_{\beta}}\times\nwl
&\suml_{\beta^\prime\neq\beta} \frac{S_{\beta,\beta^\prime}\frac{\sigma^\beta}{n_{\pi^{-1}(\beta)} m_{\beta}}}
{\sqrt{\left(\frac{\sigma^\beta}{n_{\pi^{-1}(\beta)} m_{\beta}}\right)^2
+
\left(\frac{\sigma^{\beta^\prime}}{n_{\pi^{-1}(\beta^\prime)} m_{\beta^\prime}}\right)^2}}
\nwl
&\Lambda_{\alpha,\beta}=\biggl(T_\alpha+U^\beta+\nwl
&\frac 1{n_\alpha m_\beta}\frac{R_{\alpha,\pi^{-1}(\beta)}\frac{\sigma^\beta}{n_{\pi^{-1}(\beta)} m_{\beta}}+S_{\beta,\pi(\alpha)}\frac{\sigma_\alpha}{n_{\alpha} m_{\pi(\alpha)}}}{\sqrt{\left(\frac{\sigma_\alpha}{n_{\alpha} m_{\pi(\alpha)}}\right)^2
+
\left(\frac{\sigma^{\beta^\prime}}{n_{\pi^{-1}(\beta^\prime)} m_{\beta^\prime}}\right)^2}}\biggr)^{-1}
\end{eqnarray}

Then, we have the following more general result:
\begin{theorem}\label{thm:ext}
\
\begin{enumerate}
\item Suppose that $\tlD_{\alpha,\alpha^\prime}=  D_{\alpha,\pi(\alpha^\prime)}$ satisfies the strong cyclical monotonicity condition, where for each
simple loop $i_1\to i_2\to \ldots \to i_k\to i_{k+1}=i_1$ with length $k>1$ we have
\begin{equation}\label{eq:strong_pos_loop_cond_kernel}
\suml_{l=1}^k\tlD_{i_li_{l+1}}\geq \suml_{l=1}^k\tlD_{i_li_{l}}+k\delta.
\end{equation}
Then, the solution ${X_{\alpha,\beta}}$ of the characteristic optimization in \eqref{eq:char_kernel} satisfies one of the following:
\begin{enumerate}
    \item If $\bdelta=0$ and $\theta=\infty$, we have $X_{\alpha,\beta}=\delta_{\beta,\pi(\alpha)}\frac{\sigma_\alpha}{m_\alpha n_\beta}$ with $\delta_{\ldotp,\ldotp}$ being the Kronecker index, if
    \[
    \lambda\leq\delta\maxl_{\alpha\neq\beta}\Lambda_{\alpha,\beta}
    \]
    \item Otherwise, 
\[
\delta\suml_{\beta\neq \pi(\alpha)}X_{\alpha,\beta}\leq
\]
\[
\lambda\suml_{\alpha\neq\alpha^\prime}
\biggl(\frac{R_{\alpha,\alpha^\prime}}{n_\alpha n_{\alpha^\prime}}
\sqrt{\frac{n_{\alpha^\prime}^2\sigma^2_\alpha}{m_{\pi(\alpha)}}+
\frac{n_{\alpha}^2\sigma^2_{\alpha^\prime}}{m_{\pi(\alpha^\prime)}}
}\]
\[
+\frac{S_{\pi(\alpha),\pi(\alpha^\prime)}}{m_{\pi(\alpha)} m_{\pi(\alpha^\prime)}}
\sqrt{\frac{m^2_{\pi(\alpha^\prime)}\sigma^2_\alpha}{n_\alpha}+
\frac{m^2_{\pi(\alpha)}\sigma^2_{\alpha^\prime}}{n_{\alpha^\prime}}
}\biggr)
\]
\[
+\frac\theta 2\left(\suml_\alpha \frac{\delta_\alpha^2}{n_\alpha}+\suml_\alpha \frac{\delta_\alpha^2}{m_{\pi(\alpha)}}\right)
+\frac {\Delta_1^2n}{\theta}
+\Delta_0\left(\|\bdelta\|_1-\|\bdelta\|_\infty\right)
\]
where
\[
\Delta_0=\max_{\alpha,\alpha^\prime}\left|2\tlD_{\alpha,\alpha^\prime}-
\tlD_{\alpha,\alpha}-\tlD_{\alpha^\prime,\alpha^\prime}\right|,
\]
\[
\quad\Delta_1=\frac{\Delta_0+\maxl_{\alpha} |\tlD_{\alpha,\alpha}|}2
\]

\end{enumerate}
\item The solution of \eqref{eq:main_kernel} is given by $X_{ij}=X_{\alpha,\beta}$ if there exist positive constants $a,c,d$ such that $2a+c+d\leq 1$ and for all $i,i^\prime\in C_\alpha$ and $j,j^\prime\in D_\beta$,
\[
\sqrt{\suml_{j\in [m]}\left(D_{ij}-D_{i^\prime j}\right)^2}\leq 2a n_\alpha\lambda R_{i,i^\prime}
,\]
\[
\quad
\sqrt{\suml_{i\in [n]}\left(D_{ij}-D_{i j^\prime}\right)^2}\leq 2a m_\beta\lambda S_{j,j^\prime}
\]
\[
|\mu_i-\mu_{i^\prime}|\leq\frac{c\lambda n_\alpha R_{i,i^\prime}}{\theta\sqrt{m}}
,\quad|\nu_j-\nu_{j^\prime}|\leq\frac{c\lambda m_\beta S_{j,j^\prime}}{\theta\sqrt{n}}
\]
\[
\sqrt{\left(\suml_{\alpha^\prime\neq\alpha}
\frac{R_{i,\alpha^\prime}-R_{i^\prime,\alpha}}{\sqrt{m_\alpha+m_{\alpha^\prime}}}
\right)^2
+
\suml_{\alpha^\prime\neq\alpha}\left(
\frac{R_{i,\alpha^\prime}-R_{i^\prime,\alpha}}{\sqrt{m_\alpha+m_{\alpha^\prime}}}
\right)^2
}
\leq
\]
\[
dn_\alpha R_{i,i^\prime}
\]
\[
\sqrt{\left(\suml_{\beta^\prime\neq\beta}
\frac{S_{j,\beta^\prime}-S_{j^\prime,\beta}}{\sqrt{n_\beta+n_{\beta^\prime}}}
\right)^2
+
\suml_{\alpha^\prime\neq\alpha}\left(
\frac{S_{j,\beta^\prime}-S_{j^\prime,\beta}}{\sqrt{n_\beta+n_{\beta^\prime}}}
\right)^2
}
\leq 
\]
\[
dm_\beta S_{j,j^\prime}
\]

\end{enumerate}
\begin{proof}
Denote the optimal value of \eqref{eq:ideal_kernel} and \eqref{eq:char_kernel} by $C_0$ and $C_1$, respectively. Also, notice that since $\tlD_{\alpha,\alpha^\prime}$ satisfies the strong cyclical monotonicity condition, $Y_{\alpha,\beta}=\delta_{\beta,\pi(\alpha)}\sigma_\alpha$ is the solution of \eqref{eq:ideal_kernel} and there exist dual variables $p_\alpha, q_\beta$ such that
\[
D_{\alpha,\beta}- p_\alpha-q_\beta\left\{\begin{array}{lc}
=0 & \beta=\pi(\alpha)\\
\geq \delta & \beta\neq\pi(\alpha)
\end{array}\right.
\]
Moreover,
\[
C_0=\suml_{\alpha}\sigma_\alpha p_\alpha+\suml_{\beta}\sigma^\beta q_\beta
\]

For part 1.a, we note that under the given conditions, the solution $X_{\alpha,\beta}$ of \eqref{eq:char_kernel} coincides with that of \eqref{eq:char_kernel_tight}. Now, we show that $X^\prime_{\alpha,\beta}=\frac{Y_{\alpha,\beta}}{n_\alpha m_\beta}=\frac{\delta_{\beta,\pi(\alpha)}\sigma_\alpha}{n_\alpha m_\beta}$ satisfies with the dual parameters $p^\prime_\alpha,q^\prime_\beta$, the optimality condition of \eqref{eq:char_kernel_tight}, which can be written as
\[
(D_{\alpha,\beta}- p^\prime_\alpha-q^\prime_\beta)n_\alpha m_\beta+\lambda A_{\alpha,\beta}\left\{\begin{array}{lc}
=0 & \beta=\pi(\alpha)\\
\geq 0 & \beta\neq\pi(\alpha)
\end{array}\right.
\]
where $A_{\alpha,\beta}$ is the partial derivative at $X^\prime_{\alpha,\beta}$ of the SON term w.r.t $X_{\alpha,\beta}$. By direct calculation, we observe that 
\[
A_{\alpha,\beta}=\left\{\begin{array}{lc}
n_\alpha m_\beta (T_\alpha+U^\beta)
& \beta=\pi(\alpha)\\
-\frac{R_{\alpha,\pi^{-1}(\beta)}\frac{\sigma^\beta}{n_{\pi^{-1}(\beta)} m_{\beta}}+S_{\beta,\pi(\alpha)}\frac{\sigma_\alpha}{n_{\alpha} m_{\pi(\alpha)}}}{\sqrt{\left(\frac{\sigma_\alpha}{n_{\alpha} m_{\pi(\alpha)}}\right)^2
+
\left(\frac{\sigma^{\beta^\prime}}{n_{\pi^{-1}(\beta^\prime)} m_{\beta^\prime}}\right)^2}} & \beta\neq\pi(\alpha)
\end{array}\right.
\]
It is now simple to check that under the given assumption, taking $p_\alpha^\prime=p_\alpha+\lambda T_\alpha$ and $q_\beta^\prime=q_\beta+\lambda U^\beta$ will satisfy the optimality conditions.

For part 1.b, we note that for the solution $X_{\alpha,\beta}$ of \eqref{eq:char_kernel},
\[
C_1=F\left(\{X_{\alpha,\beta}\}\right)\geq\suml_{\alpha,\beta}n_\alpha m_\beta X_{\alpha,\beta}D_{\alpha,\beta}
+\]
\[
\frac\theta 2\left(\suml_\alpha n_\alpha\left(\ba_M^T\bx_\alpha-\mu_\alpha\right)^2
+\suml_\beta m_\beta\left(\ba_N^T\bx^\beta-\nu^\beta\right)^2\right)
\]
\[
=\suml_{\alpha,\beta} n_\alpha m_\beta X_{\alpha,\beta}\left( D_{\alpha,\beta}-p_\alpha-q_\beta\right)+\suml_\alpha p_\alpha\sigma_\alpha+\suml_{\beta}\sigma^\beta q_\beta
\]
\[
+\suml_\alpha(\ba_M^T\bx_\alpha-\mu_\alpha)p_\alpha n_\alpha+
\suml_\beta(\ba_N^T\bx^\beta-\nu^\beta)q_\beta m_\beta+
\]
\[
\suml_\alpha(\mu_\alpha n_\alpha-\sigma_\alpha)p_\alpha+
\suml_\beta(\nu^\beta m_\beta-\sigma^\beta)q_\beta
\]
\[
+\frac\theta 2\left(\suml_\alpha n_\alpha\left(\ba_M^T\bx_\alpha-\mu_\alpha\right)^2
+\suml_\beta m_\beta\left(\ba_N^T\bx^\beta-\nu^\beta\right)^2\right)
\]
\[
\geq\delta\suml_{\beta\neq\pi(\alpha)}X_{\alpha,\beta}+C_0+\suml_\alpha p_\alpha\delta_\alpha+\suml_{\beta}\delta^\beta q_\beta
\]
\[
-\frac 1{2\theta}\left(\suml_\alpha p_\alpha^2n_\alpha+
\suml_\beta q_\beta^2m_\beta\right),
\]

where $F(\ldotp)$ denotes the objective function in \eqref{eq:char_kernel}. On the other hand for $X^\prime_{\alpha,\beta}=\frac{Y_{\alpha,\beta}}{n_\alpha m_\beta}=\frac{\delta_{\beta,\pi(\alpha)}\sigma_\alpha}{n_\alpha m_\beta}$, we have that
\[
C_1\leq F(\{X^\prime_{\alpha,\beta}\})= C_0
+
\]
\[
\lambda\suml_{\alpha\neq\alpha^\prime}
\biggl(\frac{R_{\alpha,\alpha^\prime}}{n_\alpha n_{\alpha^\prime}}
\sqrt{\frac{n_{\alpha^\prime}^2\sigma^2_\alpha}{m_{\pi(\alpha)}}+
\frac{n_{\alpha}^2\sigma^2_{\alpha^\prime}}{m_{\pi(\alpha^\prime)}}
}
+\]
\[
\frac{S_{\pi(\alpha),\pi(\alpha^\prime)}}{m_{\pi(\alpha)} m_{\pi(\alpha^\prime)}}
\sqrt{\frac{m^2_{\pi(\alpha^\prime)}\sigma^2_\alpha}{n_\alpha}+
\frac{m^2_{\pi(\alpha)}\sigma^2_{\alpha^\prime}}{n_{\alpha^\prime}}
}\biggr)
\]
\[
+\frac\theta 2\left(\suml_\alpha \frac{\delta_\alpha^2}{n_\alpha}+\suml_\alpha \frac{\delta_\alpha^2}{m_{\pi(\alpha)}}\right)
\]
We conclude that
\[
\delta\suml_{\beta\neq\pi(\alpha)}X_{\alpha,\beta}\leq
\]
\[
\lambda\suml_{\alpha\neq\alpha^\prime}
\biggl(\frac{R_{\alpha,\alpha^\prime}}{n_\alpha n_{\alpha^\prime}}
\sqrt{\frac{n_{\alpha^\prime}^2\sigma^2_\alpha}{m_{\pi(\alpha)}}+
\frac{n_{\alpha}^2\sigma^2_{\alpha^\prime}}{m_{\pi(\alpha^\prime)}}
}
\]
\[
+
\frac{S_{\pi(\alpha),\pi(\alpha^\prime)}}{m_{\pi(\alpha)} m_{\pi(\alpha^\prime)}}
\sqrt{\frac{m^2_{\pi(\alpha^\prime)}\sigma^2_\alpha}{n_\alpha}+
\frac{m^2_{\pi(\alpha)}\sigma^2_{\alpha^\prime}}{n_{\alpha^\prime}}
}\biggr)
\]
\[
+\frac\theta 2\left(\suml_\alpha \frac{\delta_\alpha^2}{n_\alpha}+\suml_\alpha \frac{\delta_\alpha^2}{m_{\pi(\alpha)}}\right)+
\frac 1{2\theta}\left(\suml_\alpha p_\alpha^2n_\alpha+
\suml_\beta q_\beta^2m_\beta\right)
-\]
\[
\suml_\alpha p_\alpha\delta_\alpha-\suml_{\beta}\delta^\beta q_\beta
\]
Lemma \ref{lem} gives the result in part 1.

For part 2, notice that the optimality condition of $X_{\alpha,\beta}$ yields
\[
n_\alpha m_\beta D_{\alpha,\beta}
+\lambda\suml_{\alpha^\prime\neq\alpha}R_{\alpha,\alpha^\prime} m_\beta(\bz_{\alpha,\alpha^\prime})_\beta+
\lambda\suml_{\beta^\prime\neq\beta}S_{\beta, \beta^\prime} n_\alpha(\bz^{\beta,\beta^\prime})_\alpha
\]
\[
+\theta n_\alpha m_\beta(\ba_M^T\bx_\alpha-\mu_\alpha)+\theta m_\beta n_\alpha (\ba_N^T\bx^\beta-\nu^\beta)=0
\]
where
\[
\bz_{\alpha,\alpha^\prime}=\frac{\bx_\alpha-\bx_{\alpha^\prime}}{\|\bx_\alpha-\bx_{\alpha^\prime}\|_M}
,\ \bz^{\beta,\beta^\prime}=\frac{\bx^\beta-\bx^{\beta^\prime}}
{\|\bx^\beta-\bx^{\beta^\prime}\|_N}
\]
Define for $i,i^\prime\in C_\alpha$ and $j,j^\prime\in D_\beta$
\[
(\bz_{i,i^\prime})_j=\frac{1}{2\lambda n_\alpha R_{i,i^\prime}}\biggl(
-D_{ij}+D_{i^\prime j}
-\frac{\suml_{j^\prime\in D_\beta}D_{ij^\prime}}{m_\beta}
+
\]
\[
\frac{\suml_{j^\prime\in D_\beta}D_{i^\prime j^\prime}}{m_\beta}-
2\theta\mu_i+2\theta\mu_{i^\prime}\biggr)
\]
\[
-\frac 1{n_\alpha R_{i,i^\prime}}
\suml_{\alpha^\prime\neq\alpha}
\left(
R_{i,\alpha^\prime}-R_{i^\prime,\alpha^\prime}
\right)(\bz_{\alpha,\alpha^\prime})_\beta
\]
\[
(\bz^{j,j^\prime})_i=\frac{1}{2\lambda m_\beta S_{j,j^\prime}}
\biggl(
-D_{ij}+D_{i j^\prime}
-\frac{\suml_{i^\prime\in C_\alpha}D_{i^\prime j}}{n_\alpha}
\]
\[
+
\frac{\suml_{i^\prime\in C_\alpha}D_{i^\prime j^\prime}}{n_\alpha}
-2\theta\nu_j+2\theta\nu_{j^\prime}
\biggr)
\]
\[
-\frac 1{m_\beta S_{j,j^\prime}}
\suml_{\beta^\prime\neq\beta}
\left(
S_{j,\beta^\prime}-S_{j^\prime,\beta^\prime}
\right)(\bz^{\beta,\beta^\prime})_\alpha
\]

Also for $i\in C_\alpha, i^\prime\in C_{\alpha^\prime}$ and $j\in D_\beta, j^\prime\in D_{\beta^\prime}$, where $\alpha\neq\alpha^\prime$ and $\beta\neq\beta^\prime$, take $(\bz_{ii^\prime})_j=(\bz_{\alpha,\alpha^\prime})_\beta$, $(\bz^{jj^\prime})_i=(\bz^{\beta,\beta^\prime})_\alpha$. Then, it simple to check that $X_{ij}=X_{\alpha,\beta}$ satisfies the optimality conditions of \eqref{eq:main_kernel} under conditions of the theorem and noticing that by the root-means-square and arithmetic mean (RMS-AM) inequality, we also have
\[
\sqrt{\suml_{\beta\in [K]}{m_\beta\left(\frac{\suml_{j\in D_\beta} (D_{ij}-D_{i^\prime,j})}{m_\beta}\right)^2}}\leq 2a\lambda n_\alpha R_{i,i^\prime}
\]
\[
\sqrt{\suml_{\alpha\in [K]}{n_\alpha\left(\frac{\suml_{i\in C_\alpha} (D_{ij}-D_{ij^\prime})}{n_\alpha}\right)^2}}\leq 2a\lambda m_\beta S_{j,j^\prime}
\]
\end{proof}
\end{theorem}

\begin{lemma}\label{lem}
Suppose that the ideal optimization in \eqref{eq:ideal_kernel} has a solution where $X_{\alpha,\pi(\alpha)}>0$ holds for every $\alpha$. For every $\bdelta=(\delta_\alpha)_\alpha$ satisfying $\bone^T\bdelta=0$ and any choice of the optimal dual parameters $\{p_\alpha, q_\beta\}$ we have that
\[
\suml_\alpha p_\alpha\delta_\alpha+\suml_\beta q_\beta\delta^\beta\leq\Delta_0\left(\|\bdelta\|_1-\|\bdelta\|_\infty\right)
\]
where $\delta^\beta=-\delta_{\pi^{-1}(\beta)}$. As a result in this case, \eqref{eq:ideal_kernel} has optimal dual parameters $\{p_\alpha, q_\beta\}$ satisfying
\[
|p_\alpha|\leq \Delta_1,\ |q_\beta|\leq \Delta_1
\]
\begin{proof}
  Denote the minimum value of $X_{\alpha,\pi(\alpha)}$ by $\epsilon$. Without loss of generality, we assume that $\|\bdelta\|_1-\|\bdelta\|_\infty\leq\epsilon$. Take $\alpha_0\in\arg\minl_\alpha|\delta_\alpha|$. Hence,  $\|\bdelta\|_1-\|\bdelta\|_\infty=\suml_{\alpha\neq\alpha_0}|\delta_\alpha|$.

  Denote the optimal value of \eqref{eq:ideal_kernel} by $C_0$. From the strong duality theorem we have that
  \[
  C_0=\suml_\alpha p_\alpha\sigma_\alpha+\suml_\beta q_\beta\sigma^\beta
  \]
  Take
 \begin{eqnarray}\label{eq:ideal_perturbed_kernel}
&C_1=\minl_{Y_{\alpha,\beta}\geq 0}\suml_{\alpha,\beta}Y_{\alpha,\beta}D_{\alpha,\beta}
\nwl
&\mathrm{s.t}
\nwl
&\bone^T\by^\beta=\sigma^\beta+\delta^\beta, \bone^T\by_\alpha=\sigma_\alpha+\delta_\alpha
\end{eqnarray}
We notice that $\{p_\alpha, q_\beta\}$ are feasible dual vectors for \eqref{eq:ideal_perturbed_kernel}. Hence, from the weak duality theorem we have
\[
C_1\geq \suml_\alpha p_\alpha(\sigma_\alpha+\delta_\alpha)+\suml_\beta q_\beta(\sigma^\beta+\delta^\beta)
\]
\[
=C_0+\suml_\alpha p_\alpha\delta_\alpha+\suml_\beta q_\beta\delta^\beta
\]
Now take the solution
\[
Y^\prime_{\alpha,\beta}=Y_{\alpha,\beta}\left\{
\begin{array}{cc}
  -|\delta_\alpha| & \alpha\neq\alpha_0,\ \beta=\pi(\alpha) \\
  -\suml_{\alpha\neq\alpha_0}|\delta_\alpha| & \alpha=\alpha_0,\ \beta=\pi(\alpha_0) \\
  +(\delta^\beta)_+ & \alpha=\alpha_0,\ \beta\neq\pi(\alpha_0) \\
  +(\delta_\alpha)_+ & \alpha\neq\alpha_0,\ \beta=\pi(\alpha_0) \\
  +0 & \mathrm{Otherwise}
\end{array}
\right.
\]
It is simple to check that $Y^\prime_{\alpha,\beta}$ is feasible in \eqref{eq:ideal_perturbed_kernel}. Moreover, we have
\[
C_1\leq \suml_{\alpha,\beta}Y^\prime_{\alpha,\beta}D_{\alpha,\beta}=C_0+
\]
\[
\suml_{\alpha\neq \alpha_0}
\left(
2D_{\alpha_0\pi(\alpha)}(\delta_\alpha)_++
2D_{\alpha\alpha_0}(\delta_\alpha)_--
\right.
\]
\[
\left.
(D_{\alpha,\alpha}+D_{\alpha_0,\alpha_0})|\delta_\alpha|
\right)
\]
\[
\leq C_0+\Delta_0\suml_{\alpha\neq\alpha_0}|\delta_\alpha|
\]
We conclude that
\[
\suml_\alpha p_\alpha\delta_\alpha+\suml_\beta q_\beta\delta^\beta\leq \Delta_0\suml_{\alpha\neq\alpha_0}|\delta_\alpha|
\]
which proves the first part. Now, notice that for any pair $(\alpha_1,\alpha_2)$ of distinct indices, taking $\delta_{\alpha_1}=1$ and $\delta_{\alpha_1}=-1$ gives
\[
p_{\alpha_1}-p_{\alpha_2}-q_{\alpha_1}+q_{\alpha_2}\leq\Delta_0
\]
switching $\alpha_1,\alpha_2$ yield
\[
|p_{\alpha_1}-p_{\alpha_2}-q_{\alpha_1}+q_{\alpha_2}|\leq\Delta_0
\]
Now, notice that from the optimality of \eqref{eq:ideal_kernel} we have $p_\alpha+q_\alpha=D_{\alpha,\alpha}$, which leads to
\[
2|p_{\alpha_1}-p_{\alpha_2}|\leq\Delta_0+|D_{\alpha_1,\alpha_1}-D_{\alpha_2,\alpha_2}|
\]
which yield
\[
\left|
\left(
p_{\alpha_1}+\frac{D_{\alpha_1,\alpha_1}}2\right)
-
\left(p_{\alpha_2}+\frac{D_{\alpha_2,\alpha_2}}2\right)
\right|\leq\Delta_0
\]
The result is obtained by noticing that the set of optimal dual solutions is invariant under shift, i.e. $p_i+\lambda$ and $q_i-\lambda$ are also solutions for any $\lambda\in\bbR$. Hence, we may take $\lambda$ such that
\[
\left|p_\alpha+\frac{D_{\alpha,\alpha}}2\right|\leq \frac{\Delta_0}2
\]
and hence
\[
\left|q_\alpha-\frac{D_{\alpha,\alpha}}2\right|\leq \frac{\Delta_0}2
\]
Triangle inequality gives the result.

\end{proof}

\end{lemma}
\subsection{Proof of Theorem \ref{thm:guarantee}}
\label{proof:guarantee}
The first claim that $X_{ij}=X_{\alpha,\beta}$ follows by specializing part 2 of Theorem \ref{thm:ext} for the conditions of Theorem \ref{thm:guarantee} with $a=1/2, b=c=0$: $\theta=\infty$ and inside clusters we have $R_{i,i^\prime}=S_{j,j^\prime=1}=1$,  $R_{i,\alpha^\prime}=R_{i^\prime,\alpha}$ and $S_{j,\beta^\prime}=R_{j^\prime,\beta}$. Moreover $n_\alpha=m_\beta=m$. 

Part 1 in the main text also is achieved by specializing part 1.a.: We will have $\sigma_\alpha=\omega_\alpha$, $n_\alpha=m_\beta=m$, $R_{\alpha,\alpha^\prime}=m^2 R$ and $S_{\beta,\beta^\prime}=m^2 $. 

Finally, part 2 is a result of 1.b. with $\theta=\infty,\bdelta=\bzero$. 

\subsection{Proof of Theorem \ref{thm:simp}}
 Based on theorem \ref{thm:guarantee}, we present a sketch of the proof for theorem \ref{thm:simp}. Under the assumptions of theorem \ref{thm:simp}, we directly verify that $\Lambda=\Lambda_{\alpha,\beta}=\nicefrac{\sqrt 2}{K(1+R)}$. For part 1) $\delta=D-d-2\omega,\ \Delta\leq 2\omega$ and for part 2) $\delta=D^2-d^2$ are valid choices. Finally, for part 2) we may conclude by Chernoff bound that with a probability exceeding $1-\nicefrac 1{n^{10}}$ (the power 10 is arbitrary) we have $\Delta=O(\omega\sqrt{(E+\omega)^2+1}\log(nK))$. Replacing these expression in the first part of theorem \ref{thm:guarantee} gives us the result.

\section{Proof of Theorem \ref{thm:convergence}}
\label{proof:convergence}
To simplify the notation, we introduce $\phi_{P+q}=I_{S_q}$ for $q=1,2\ldots, Q$, where $I_S$ denotes the indicator function of a convex set $S$. It is well-known that the proximal operator of $I_S$ coincides with the orthogonal projection operator onto $S$. Hence, we may simplify our algorithm to 
\begin{equation*}
    \bx_{t+1}=\mathrm{prox}\left(\bx_t+\mu\bg_{r_t}\right),\quad \ba_t=\rho\frac{\bx_t-\bx_{t+1}}\mu-\alpha\barbg_t, 
\end{equation*}
\begin{equation*}
    \bg_{r_t}\gets \bg_{r_t}+\ba_t,
\end{equation*}
where we introduce $\bg_{P+q}=\bh_q$ for $q=1,2\ldots, Q$ and denote $\barbg_t=\suml_{r=0}^R\bg_r$ with $R=P+Q$. Moreover, $r_t$ is equal to either $p_t$ or $P+q_t$, depending on the random choice. We also define $\bx^\dagger_{r,t}=\mathrm{prox}\left(\bx_t+\mu\bg_{r}\right)$ and hence $\bx_{t+1}=\bx^\dagger_{r,t}$.

To prove convergence, we adopt a so-called Lyaponov function approach. We introduce two non-negative functions $L,M$ of the state variables, $\bx$ and $\{\bg_r\}$ such that
\begin{equation}\label{eq:lyaponov}
    \bbE[L_{t+1}]-\bbE[L_{t}]+\bbE[M_t]\leq 0,\quad t=1,2,\ldots,
\end{equation}
where $L_t,M_t$ denote the values of $L,M$ at the variables of the $t^\tth$ iteration. Then, summing these inequalities up to an arbitrary time $t$ gives
\begin{equation}
    \bbE[L_{t}]-L_{0}+\suml_{\tau=0}^{t-1}\bbE[M_\tau]\leq 0
\end{equation}
which by the non-negativity of $L$ implies
\begin{equation}\label{eq:final_bound}
    \suml_{\tau=0}^{t-1}\bbE[M_\tau]\leq L_0.
\end{equation}
In particular, we take 
\begin{eqnarray}
    &M_t=F_t+\frac{1-\rho} {2\mu(1+\rho)}\suml_{r}\left\|\bx_t-\bx_{r,t}^\dagger\right\|^2\nwl
   &+\frac{\mu}{\rho}\left[\frac{2+\alpha}{2}
-\frac{\alpha^2}{1-\rho}
\right]
\left\|\barbg_t\right\|^2
\end{eqnarray}
where
\begin{equation}\label{eq:defF}
    F_t=\suml_{r=1}^P\left[\phi_r\left(\bx^\dagger_{r,t}\right)-\phi_r(\bx^*)-\langle\bg^*_r,\bx^\dagger_{r,t}-\bx^*\rangle\right]
\end{equation}
and $\bg_r^*\in\partial \phi_r(\bx^*)$ for $r\in [R]$ satisfy the monotone inclusion problem in \eqref{eq:first_order} at $\bx^*$, i.e $\suml_r\bg_r^*=0$. Then, the non-negativity of each summand of $F_t$ follows from the convexity of $\phi_r$. The third term of $M_t$ is also positive for $\alpha<\frac{1+\sqrt{17}}4(1-\rho)$. This establishes the non-negativity of $M_t$. We further define
\begin{equation}
\begin{aligned}
&\Gamma_t=\suml_{r=1}^R\|\bg_r-\bg^*_r\|^2,\quad
G_t=\|\mu\barbg_t+\rho(\bx_t-\bx^*)\|^2
,\\
&\quad
D_t=\left\|\bx^*-\bx_t\right\|^2
\end{aligned}
\end{equation}
and take
\begin{equation}
    L_t=\frac R {2\mu} D_t+\frac 1 {2\rho\mu\alpha}G_t+\frac {R\mu}{2\rho}\Gamma_t
\end{equation}

\subsection{Proof of Theorem Under \eqref{eq:lyaponov}}
Let us first prove the theorem assuming \eqref{eq:lyaponov} holds true for the given $L,M$. Then \eqref{eq:final_bound} also holds true and we conclude from the definition of $M$ that
\begin{equation}
\begin{aligned}
\label{eq:middle_bounds}
    &\frac 1 t\suml_{\tau=0}^{t-1} \bbE[F_\tau]\leq\frac{L_0}t,\\
    &\quad
    \frac{1-\rho} {2t(1+\rho)}\suml_{\tau=0}^{t-1}\suml_{r}\bbE\left[\left\|\bx_t-\bx_{r,t}^\dagger\right\|^2\right]\leq\frac{\mu L_0}t
\end{aligned}
\end{equation}
Now from \eqref{eq:defF} and the triangle inequality, we conclude that
\begin{equation}
\begin{aligned}
    &\bbE[F_\tau]\geq
    \bbE\left[\suml_{r=1}^P\phi_r(\bx_\tau)-\phi_r(\bx^*)\right]\\
    -
    &\suml_{r=1}^P\bbE\left|\phi_r\left(\bx^\dagger_{r,\tau}\right)-\phi_r(\bx_\tau)\right|-\\
    &\suml_{r=1}^P\bbE\left|\langle\bg^*_r,\bx^\dagger_{r,\tau}-\bx_\tau\rangle\right|
\end{aligned}
\end{equation}
Further since the functions are $\beta-$Lipschitz, we observe that $\|\bg_r^*\|\leq\beta$. Hence,
\begin{equation}
\begin{aligned}
    &\suml_{r=1}^P\bbE\left|\langle\bg^*_r,\bx^\dagger_{r,\tau}-\bx_\tau\rangle\right|\leq\beta \suml_{r=1}^P\bbE\left\|\bx^\dagger_{r,\tau}-\bx_\tau\right\|\\
    &\leq\beta\sqrt{P}\sqrt{\suml_{r=1}^P\bbE\left\|\bx^\dagger_{r,\tau}-\bx_\tau\right\|^2}
\end{aligned}
\end{equation}
Similarly,
\begin{equation}
\begin{aligned}
    &\suml_{r=1}^P\bbE\left|\phi_r\left(\bx^\dagger_{r,\tau}\right)-\phi_r(\bx_\tau)\right|\leq 
    \beta \suml_{r=1}^P\bbE\left\|\bx^\dagger_{r,\tau}-\bx_\tau\right\|\\
    &\leq\beta\sqrt{P}\sqrt{\suml_{r=1}^P\bbE\left\|\bx^\dagger_{r,\tau}-\bx_\tau\right\|^2}
\end{aligned}
\end{equation}
We conclude that
\begin{equation}
\begin{aligned}
    &\bbE\left[\suml_{r=1}^P\phi_r(\bx_\tau)-\phi_r(\bx^*)\right]\\
    &\leq \bbE[F_\tau]+2\beta\sqrt{P}\sqrt{\suml_{r=1}^P\bbE\left\|\bx^\dagger_{r,\tau}-\bx_\tau\right\|^2}
\end{aligned}
\end{equation}
and hence,
\begin{eqnarray}
    &\frac 1 t\suml_{\tau=0}^{t-1}\bbE\left[\suml_{r=1}^P\phi_r(\bx_\tau)-\phi_r(\bx^*)\right]\nwl
    &\leq \frac 1 t\suml_{\tau=0}^{t-1}\bbE[F_\tau]+\frac {2\beta\sqrt{P}} t\suml_{\tau=0}^{t-1}\sqrt{\suml_{r=1}^P\bbE\left\|\bx^\dagger_{r,\tau}-\bx_\tau\right\|^2}\nwl
    &\leq\frac 1 t\suml_{\tau=0}^{t-1}\bbE[F_\tau]+\nwl
    &2\beta\sqrt{P}\sqrt{\frac 1 t\suml_{\tau=0}^{t-1}\suml_{r=1}^P\bbE\left\|\bx^\dagger_{r,\tau}-\bx_\tau\right\|^2}
\end{eqnarray}
Using Jensen's inequality and \eqref{eq:middle_bounds}, we obtain
\begin{equation}
\begin{aligned}
    &\bbE\left[\suml_{r=1}^P\phi_r(\barbx_\tau)-\phi_r(\bx^*)\right]\leq \\
    &\frac 1 t\suml_{\tau=0}^{t-1}\bbE\left[\suml_{r=1}^P\phi_r(\bx_\tau)-\phi_r(\bx^*)\right]\leq\\
    &\frac {L_0}t+2\beta\sqrt{\frac{2P\mu L_0(1+\rho)}{t(1-\rho)}}
\end{aligned}
\end{equation}
This proves the first bound in part 1 noting that for a suitable constant $c$ only depending on $\rho,\alpha$ and the constant in \eqref{eq:assu_1}
\begin{equation}
    L_0\leq c\beta P\lambda.
\end{equation}
For the second bound in part 1 note that for $r=P+1,P+2,\ldots,P+Q$, we have $\mathrm{dist}(\bx_t,S_r)\leq\|\bx_{r,t}^\dagger-\bx_t\|$, since by definition $\bx_{r,t}^\dagger\in S_r$. We conclude that
\begin{equation}
\suml_{q=1}^Q\mathrm{dist}^2(\barbx_t,S_q)\leq\frac 1 t\suml_{\tau=0}^{t-1}\suml_{r=P+1}^R\|\bx_{r,t}^\dagger-\bx_t\|^2\leq\frac{\mu L_0}t
\end{equation}
For part 2, note that
\begin{equation}
\begin{aligned}
&\bbE\left[\left\|\bx_{t+1}-\bx_t\right\|^2\right]=\bbE\left[\bbE\left[\left\|\bx_{t+1}-\bx_t\right\|^2\mid\bx_t\right]\right]=\nwl &\bbE\left[\frac 1 R\suml_r\left\|\bx^\dagger_{r,t}-\bx_t\right\|^2\right]
\end{aligned}
\end{equation}
We conclude from \eqref{eq:final_bound} that
\begin{equation}
\suml_t\bbE\left[\left\|\bx_{t+1}-\bx_t\right\|^2\right]\leq \frac{\mu L_0}{R}
\end{equation}
which completes the proof.
\subsection{Proof of \eqref{eq:lyaponov}}
It remains to prove \eqref{eq:lyaponov}. 
We first state the following intermediate results that characterize the term $\bbE[L_t]-\bbE[L_{t-1}]$ in \eqref{eq:lyaponov}. They can be proven by direct substitution and hence the proofs are neglected.
\begin{lemma}
 The average dynamics of $G_t$ is given by:
\begin{equation}
\begin{aligned}
\label{eq:LG}
    &\bbE[G_{t+1}]-\bbE[G_{t}]=-\mu^2\left[1-(1-\alpha)^2\right]\bbE\|\barbg_t\|^2\\
    &+2\rho\mu\alpha \bbE\left\langle
\bx^*-\bx_t,\barbg_t
\right\rangle
\end{aligned}
\end{equation}
\end{lemma}

\begin{lemma}
The average dynamics of $\Gamma_t$ is given by:
 \begin{equation}
 \begin{aligned}
 \label{eq:LGamma}
    &\bbE[\Gamma_{t+1}]-\bbE[\Gamma_{t}]=\frac{1}R
\suml_{r}
\bbE\left\|
\rho\frac{\bx_t-\bx_{r,t}^\dagger}\mu-\alpha\barbg_t
\right\|^2\\
&+\frac{2\rho}{R\mu}
\suml_{r}
\bbE\left\langle
\bx_t-\bx_{r,t}^\dagger,\bg_r-\bg_r^*
\right\rangle-\frac{2\alpha}R\|\barbg_t\|^2
 \end{aligned}
 \end{equation}
\end{lemma}

\begin{lemma}
The average dynamics of $D_t$ is given by:
\begin{equation}
\begin{aligned}
\label{eq:LD}
     &\bbE[D_{t+1}]-\bbE[D_{t}]=-\frac{1}R
\suml_{r}\bbE\|\bx^\dagger_{r,t}-\bx_t\|^2\\
&+\frac{2}R
\suml_{r}\bbE\langle \bx_t-\bx^\dagger_{r,t},\bx^*-\bx^\dagger_{r,t}\rangle
\end{aligned}
\end{equation}
\end{lemma}

Now, we state a crucial inequality that connects the dynamics of $L$ to $M$:

\begin{lemma}
The following inequality holds at every time:
\begin{equation}\label{eq:LE}
    F_t+\suml_{r=1}^R\left\langle\frac{\bx_t-\bx^\dagger_{r,t}}\mu+\bg_r-\bg_r^*,\bx^*-\bx^\dagger_{r,t}\right\rangle\leq 0
\end{equation}
\begin{proof}
From the definition of a proximal operator for $r=1,2,\ldots, P$, we have $\frac{\bx_t-\barbx^\dagger_{r,t}}{\mu}+\bg_r\in \partial\phi_r(\barbx^\dagger_{r,t})$. hence
\begin{equation}
    \phi_r(\bx^*)\geq\phi_r(\barbx^\dagger_{r,t})+\left\langle \frac{\bx_t-\barbx^\dagger_{r,t}}{\mu}+\bg_r,\bx^*-\barbx^\dagger_{r,t}\right\rangle
\end{equation}
Adding and subtracting the term $\langle\bg^*_r,\barbx^\dagger_{r,t}-\bx_t\rangle$ and summing over $r\in [P]$ gives
\begin{equation}
    F_t+\suml_{r=1}^P\left\langle\frac{\bx_t-\bx^\dagger_{r,t}}\mu+\bg_r-\bg_r^*,\bx^*-\bx^\dagger_{r,t}\right\rangle\leq 0
\end{equation}
Now, note that by the definition of a projection operator for $r=P+1,P+2,\ldots,R$ we have $\frac{\bx_t-\barbx^\dagger_{r,t}}{\mu}+\bg_r$ is normal to $S_r$ at $\barbx^\dagger_{r,t}$. Since $\bx^*\in S_r$, we have
\begin{equation}
    \left\langle\frac{\bx_t-\bx^\dagger_{r,t}}\mu+\bg_r,\bx^*-\bx^\dagger_{r,t}\right\rangle\leq 0
\end{equation}
Similarly, we obtain that 
\begin{equation}
    \langle\bg^*_r,\barbx^\dagger_{r,t}-\bx_t\rangle\leq 0
\end{equation}
Summing  (44) and (45) over $r=P+1,P+2,\ldots,R$ and adding to (43) gives the desired result.
\end{proof}
\end{lemma}

\subsubsection{Combining Bounds}
To obtain \eqref{eq:lyaponov}, we combine the inequalities in the above four lemmas in the following way. Respectively multiplying \eqref{eq:LD}, \eqref{eq:LG} and \eqref{eq:LGamma} by $\frac R{2\mu}$, $\frac 1 {2\rho\mu\alpha}$ and $\frac {R\mu}{2\rho}$ and adding to \eqref{eq:LE} and after straightforward calculations, we obtain:

\[
\bbE[L_{t+1}]-\bbE[L_{t}]+
\frac 1 {2\mu}\suml_{r}\left\|\bx_t-\bx^\dagger_{r,t}\right\|^2
+\]
\[
\mu\left[\frac{1-(1-\alpha)^2}{2\alpha\rho}+\frac\alpha\rho\right]\|\barbg_t\|^2
\]
\[
-\frac 1{2\rho\mu}
\suml_{k\in [K]}\left\|\rho\left(\bx-\bx_k^\dagger\right)-\mu\alpha\bg\right\|^2
\leq 0. \]
By invoking Jensen's inequality, we have,
\[
\left\|\rho\left(\bx-\bx_k^\dagger\right)-\mu\alpha\bg\right\|^2\leq\frac{2\rho^2}{1+\rho}\left\|\bx-\bx^\dagger_k\right\|^2+\frac{2\mu^2\alpha^2}{1-\rho}\|\bg\|^2,
\]
which yields the desired result.

\subsection{Proof of Theorem \ref{lem:proxop}}
\label{sec:proof_proxop}
The proximal operator of $\phi_{k,\bzeta,\bmeta}$ is defined as
\begin{equation}\label{eq:sthh}
\underset{(\bx,\by)\in\bbR^m\times\bbR^m}{\mathrm{argmin}}\frac 1{2\mu}\|\bx-\bp\|_2^2+
\frac 1{2\mu}\|\by-\bq\|_2^2+\phi_{\rho,\bzeta,\bmeta}(\bx,\by).
\end{equation}
We introduce a change of variables by $\bu=\nicefrac{(\bx+\by)}{2},
\bv=\nicefrac{(\bx-\by)}{2}$. First note that $\|a\|^2+\|b\|^2=(\|a+b\|^2+\|a-b\|^2)/2$. Hence
\[
\frac 1{2\mu}\|\bx-\bp\|_2^2+
\frac 1{2\mu}\|\by-\bq\|_2^2=
\]
\[
\frac 1{\mu}\left(\left\|\bu-\frac{\bp+\bq}2\right\|^2+\left\|\bv-\frac{\bp-\bq}2\right\|^2\right)
\]
Furthermore,
\[
\langle\bx,\bzeta\rangle+\langle\by,\bmeta\rangle=\langle\bu+\bv,\bzeta\rangle+\langle\bu-\bv,\bmeta\rangle=\langle\bu,\bzeta+\bmeta\rangle+\langle\bv,\bzeta-\bmeta\rangle
\]
Hence \eqref{eq:sthh} can be written as
\begin{equation}
\begin{aligned}
\label{eq:sth}
&\underset{(\bu,\bv)\in\bbR^m\times\bbR^m}{\mathrm{argmin}}\frac 1{\mu}\left(\left\|\bu-\frac{\bp+\bq}2\right\|^2+\left\|\bv-\frac{\bp-\bq}2\right\|^2\right)+\\\\
&\langle\bu,\bzeta+\bmeta\rangle+\langle\bv,\bzeta-\bmeta\rangle+2\rho\|\bv\|_2
\end{aligned}
\end{equation}
\begin{equation}
\begin{aligned}
&=\underset{(\bu,\bv)\in\bbR^m\times\bbR^m}{\mathrm{argmin}}\frac 1{\mu}\left(\left\|\bu-\frac{\bp+\bq-\mu\bzeta-\mu\bmeta}2\right\|^2\right) + \\
&\frac 1{\mu}\left(
 \left\|\bv-\frac{\bp-\bq-\mu\bzeta+\mu\bmeta}2\right\|^2\right)+2\rho\|\bv\|_2
\end{aligned}
\end{equation}

This is separable  over
$\bu$ and $\bv$, and can be analytically solved. We get 
\[
\bu=\frac{\bp+\bq-\mu\bzeta-\mu\bmeta}2,
\]
\[
\quad \bv=\calT_{\mu\rho}\left(\frac{\bp-\bq-\mu\bzeta+\mu\bmeta}2\right)
\]
The result is obtained by setting $\bx=\bu+\bv,\by=\bu-\bv$.

\section{Additional experiments}
\label{sec:additional_exp}
\subsection{Impact of SON-Regularizer}

\begin{figure*}[!th]
\centering
\includegraphics[width=0.235\textwidth]{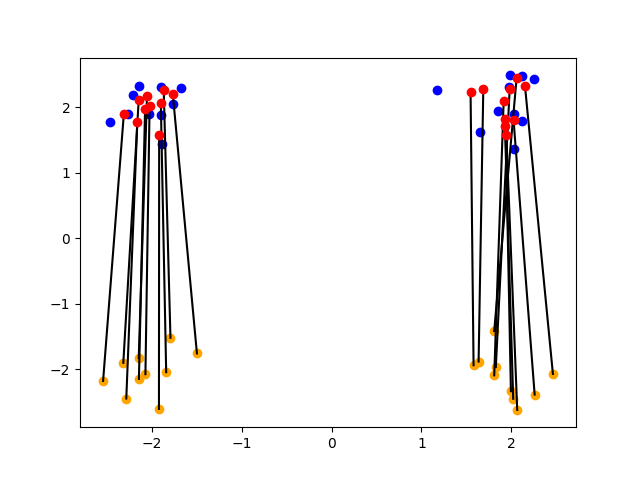}
\includegraphics[width=0.235\textwidth]{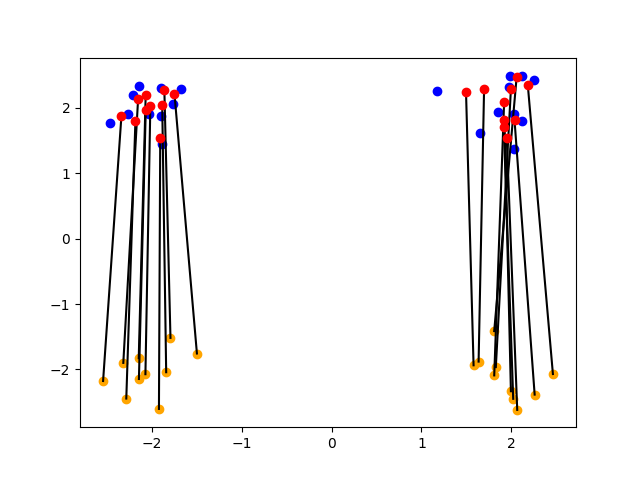}
\includegraphics[width=0.235\textwidth]{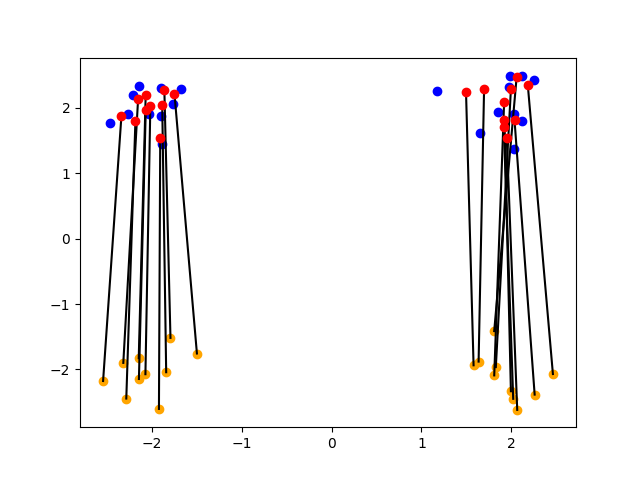}
\includegraphics[width=0.235\textwidth]{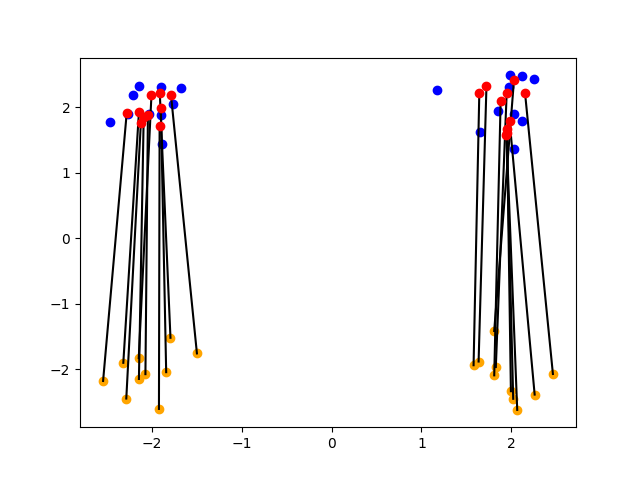}
\\
\includegraphics[width=0.235\textwidth]{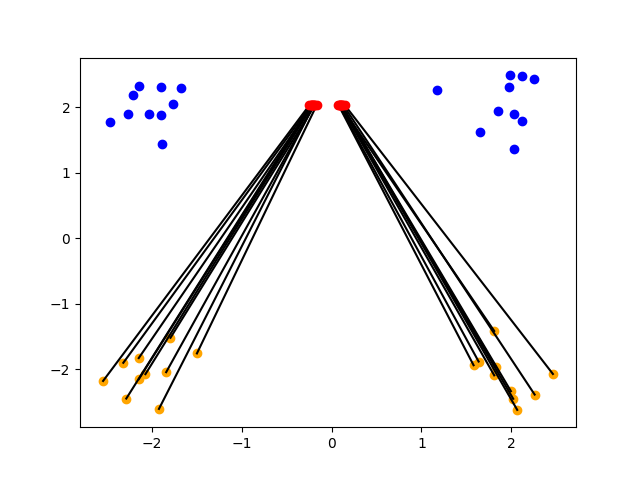}
\includegraphics[width=0.235\textwidth]{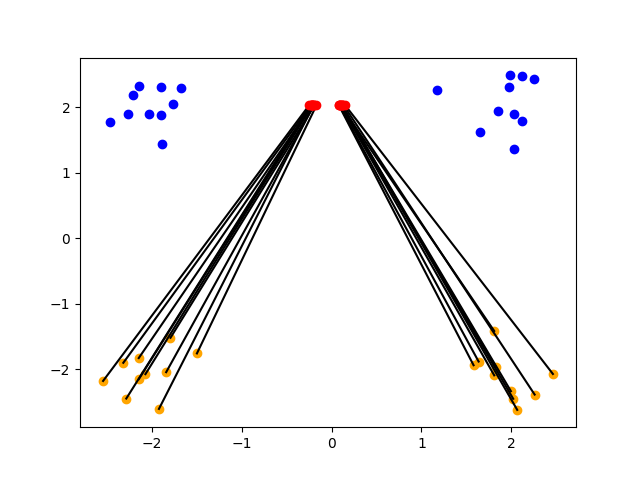}
\includegraphics[width=0.235\textwidth]{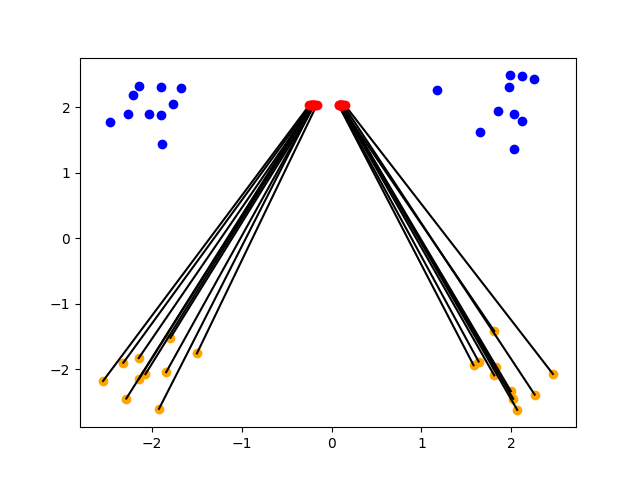}
\includegraphics[width=0.235\textwidth]{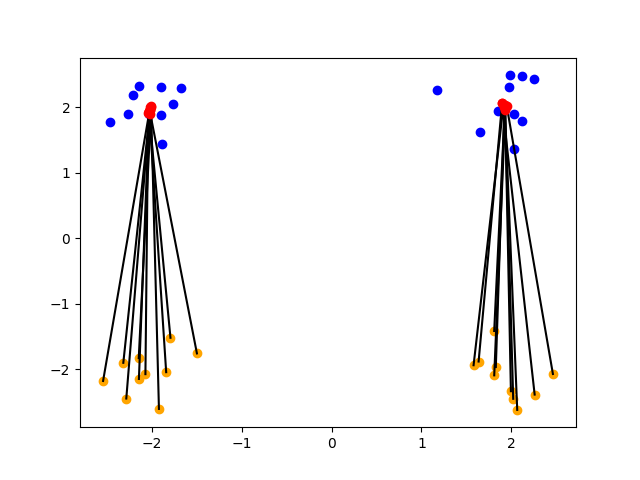}
\caption{Illustration of different models on simple data, where the source and target domains have the same number of classes and similar distributions.
The columns respectively correspond to OT-l1l2, OT-lpl1, OT-Sinkhorn and OT-SON. For each model, we illustrate the results for two different values of its regularization parameter. Among different models, OT-SON yields consistent, informative and stable transports for different regularization parameters.
}
\label{fig:simple_synthetic_data}
\end{figure*}

We investigate the models on a simple dataset, shown in Fig. \ref{fig:simple_synthetic_data}. 
We illustrate the behavior of each model with respect to two different values of its  regularization parameter (low and high) respectively at the first and the second row (low: $\lambda_1 = 0.01, \lambda_2=0.0$, high: $\lambda_1 = 10, \lambda_2 = 5$). In the low setting, instead of $\lambda_2=0.0$ any other small value also yields consistent results. The source data,  target data and  transported source data are respectively shown by yellow, blue and red points.
Each column of the sub-figures in Fig. \ref{fig:simple_synthetic_data} corresponds to a particular model performance respectively OT-l1l2, OT-lpl1, OT-Sinkhorn and OT-SON (our model). We observe that OT-SON yields stable and consistent results for different values of its parameters. Moreover, the data points transported  by the proposed model are always informative providing a good representation of the underlying classes. Whereas, the other OT models are sensitive to the values of their regularization parameters and  might thus transport the source data to somewhere in the middle of the actual target data, or away from the actual classes in the target domain.

\begin{figure*}[t!]
\centering
\includegraphics[width=1\textwidth]{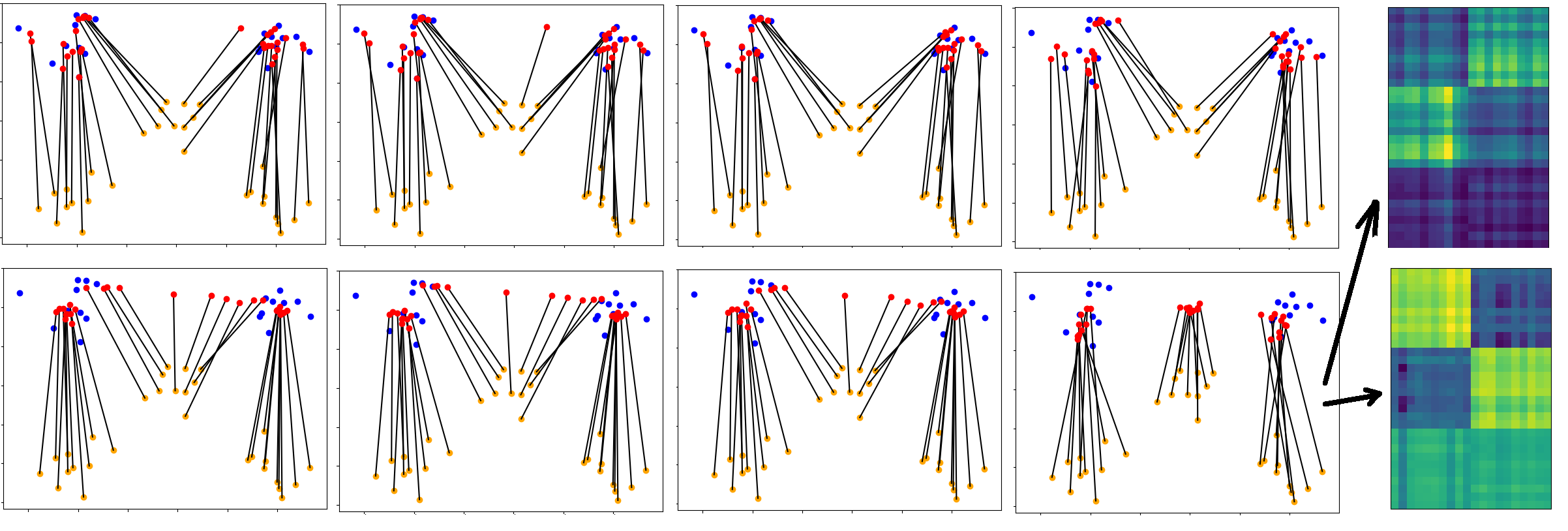}
\caption{Illustration of different methods where the source and target domains have different number of classes. 
The first four columns respectively correspond to OT-l1l2, OT-lpl1, OT-Sinkhorn and OT-SON.
Only OT-SON with an appropriate parameterization  (the forth column and the second row) identifies the presence of a superfluous class in the source and handles it properly. The last column shows the consistency between the mapping costs and the transport map.
}
\label{fig:simple_synthetic_data_diverse}
\end{figure*}

We next study the interesting case where the source and target domains do not include the same number of classes, as shown in Fig. \ref{fig:simple_synthetic_data_diverse}. In this experiment, we assume that the source data contains three classes, whereas the target domain has only two classes.
Using the same color code as in Fig.\ref{fig:simple_synthetic_data}, we see in Fig. \ref{fig:simple_synthetic_data_diverse} the target classes and the transported source classes to the target domain are shown in yellow, blue and red respectively corresponding to OT-l1l2, OT-lpl1, OT-Sinkhorn and OT-SON. We again illustrate the behavior of each model w.r.t. two different values of its  regularization parameter (low and high) respectively at the first and the second row.
We observe that among all different models, only OT-SON with an appropriate parameter is able to identify that the source and the target domains have different number of classes, and subsequently,  matches the corresponding classes correctly. It maps the superfluous class to a space between the two matched classes. However, the other models assign the superfluous class to the two other classes and do not distinguish the presence of such an extra class in the source domain. This observation is consistent with the assumptions made in \cite{CFTR17}. The unbalanced method in \cite{ChizatPSV18} might be relevant but its use is unclear to us. In the last column of Fig. \ref{fig:simple_synthetic_data_diverse}, the heat maps show the mapping cost among different source and target classes, and as well as the transport map obtained by our algorithm (OT-SON with a high regularization). We observe that the transport map respects the class structure.

\begin{figure*}[ht!]
    \centering
    \subfigure[Path-based data.]
    {
        \includegraphics[scale=0.31]{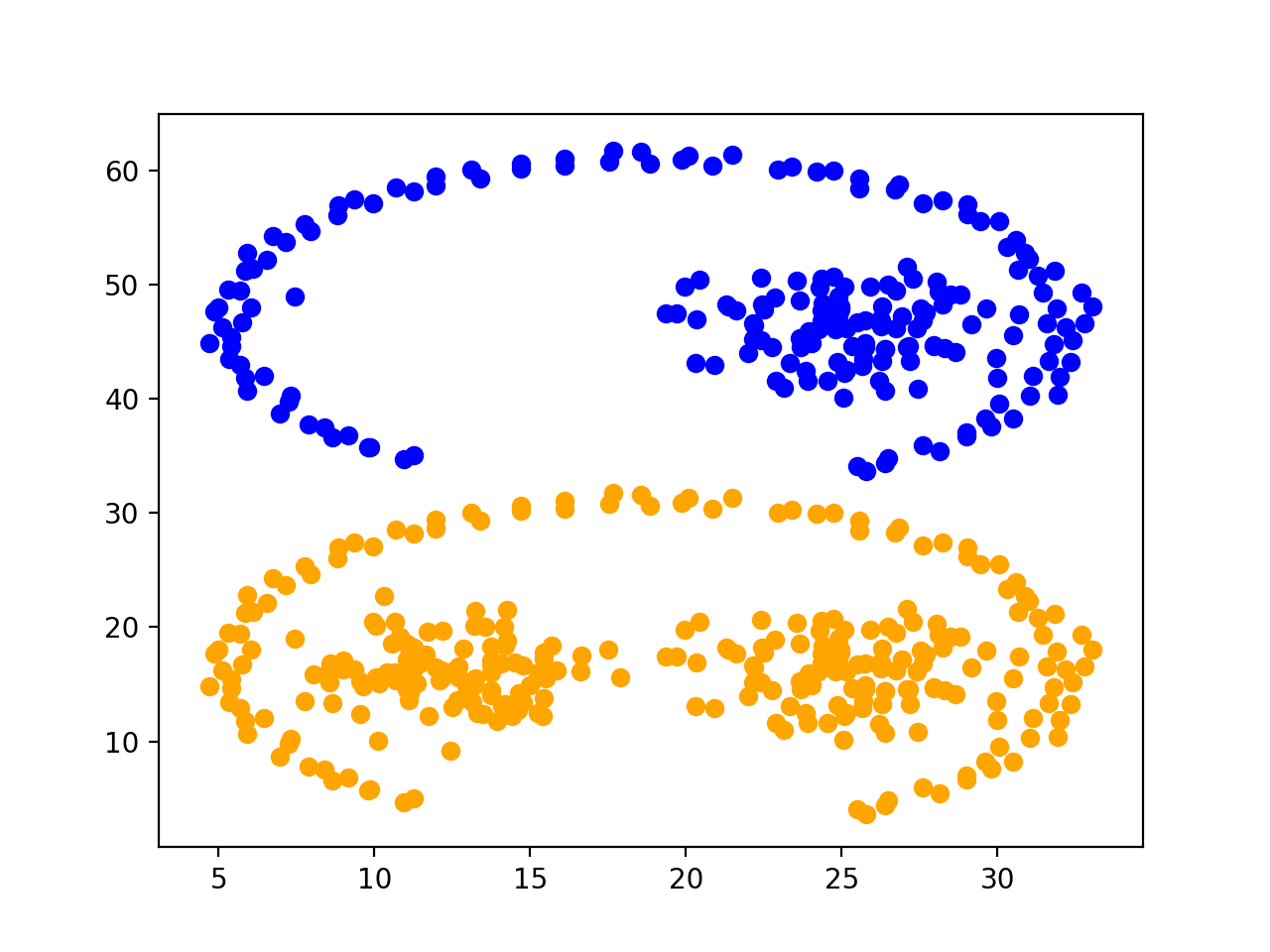}
        \label{fig:path_based_data}
    }
    \subfigure[Path-based transport.]
    {
        \includegraphics[scale=0.31]{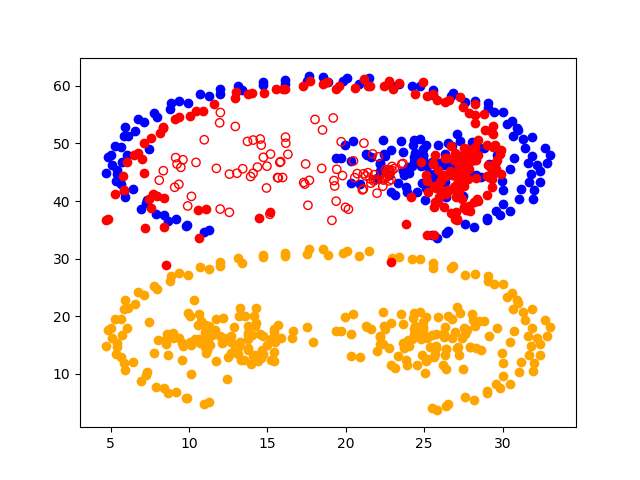}
        \label{fig:path_based_transport}
    }
    \subfigure[Accuracy results.]
    {
        \includegraphics[scale=0.17]{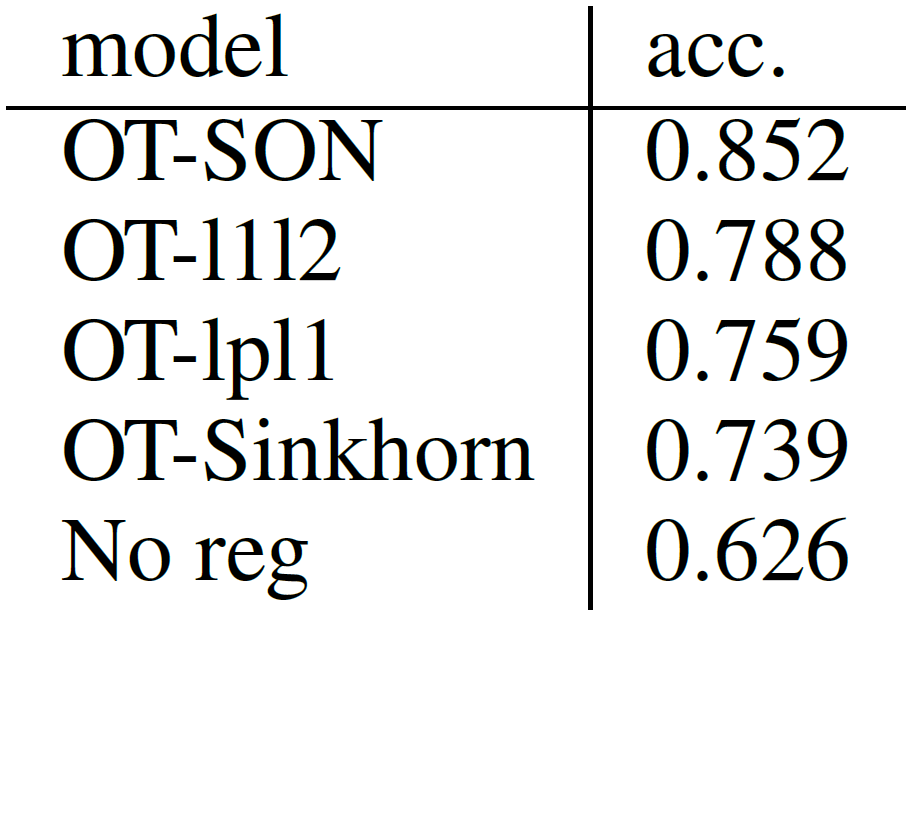}
        \label{fig:table_path}
    }
    \caption{Path-based source (yellow points) and target (blue points) datasets. 
    Using OT-SON  to transfer the path-based source data to the target domain (shown by red) yields the best results.  
}
    \label{fig:path_based}
\end{figure*}

\subsection{Experiments on path-based data}
In Fig. \ref{fig:path_based}, we investigate the different OT-based domain adaptation models on a commonly-used synthetic dataset, wherein the three  classes have diverse shapes and forms \cite{ChangY08}. 
In particular, we consider the case where one of the source classes is absent in the target domain. With the same number of classes in the source and target domains, the different models perform equally well.
Fig. \ref{fig:path_based_data} shows a case where the source data (yellow points) and the target data (blue points), differ in the fact that the target data is missing the upper left Gaussian cloud of points appearing in the source data.
Fig. \ref{fig:path_based_transport} shows the two source and target datasets, as well as the transported data by our model (OT-SON). The transported data points are shown in red. We observe that our method avoids mapping the source data of  the missing class to any of the present classes in the target domain. The points with white interior are those not assigned to any class in the target. This thus leads to a better prediction of the target data. 
In the table of Fig. \ref{fig:table_path}, we compare the accuracy scores of different models on the target data, where our model yields the highest score.



\subsection{Unsupervised domain adaptation}
\begin{figure*}[!ht]
\centering
\includegraphics[width=0.75\textwidth,,height=0.27\textwidth]{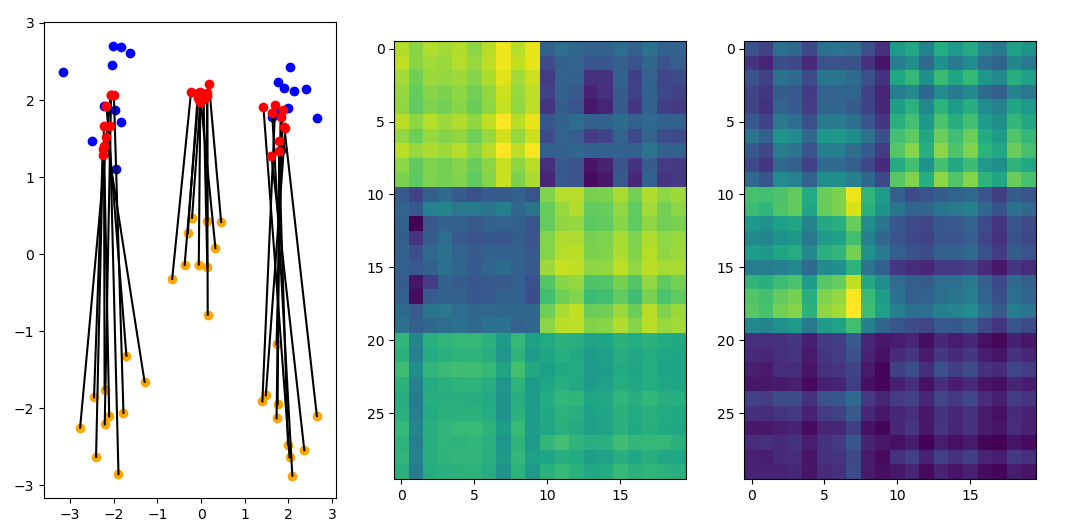}
\caption{Unsupervised OT-SON, the OT-SON results and the consistency of transport costs and transport maps. 
}
\label{fig:unsupervised_OT_SON}
\end{figure*}

In all prior experiments, we have assumed that the class labels of the source data are available. This setup is consistent with the study in \cite{CFTR17}. We consequently evaluate in a side study the fully unsupervised setting, i.e., the case where no class label is available for the source or the target data. We consider the setting used in Fig. \ref{fig:simple_synthetic_data_diverse} with, this time, no given class labels. While the other methods fail for this task, the OT-SON with proper parameterization (i.e., the setting shown in the second row and the forth column) yields meaningful and consistent results. Fig. \ref{fig:unsupervised_OT_SON} shows the OT-SON results and the consistency of transport costs and transport maps computed by OT-SON. 

\begin{figure*}[!ht]
\centering
\includegraphics[width=0.9\textwidth]{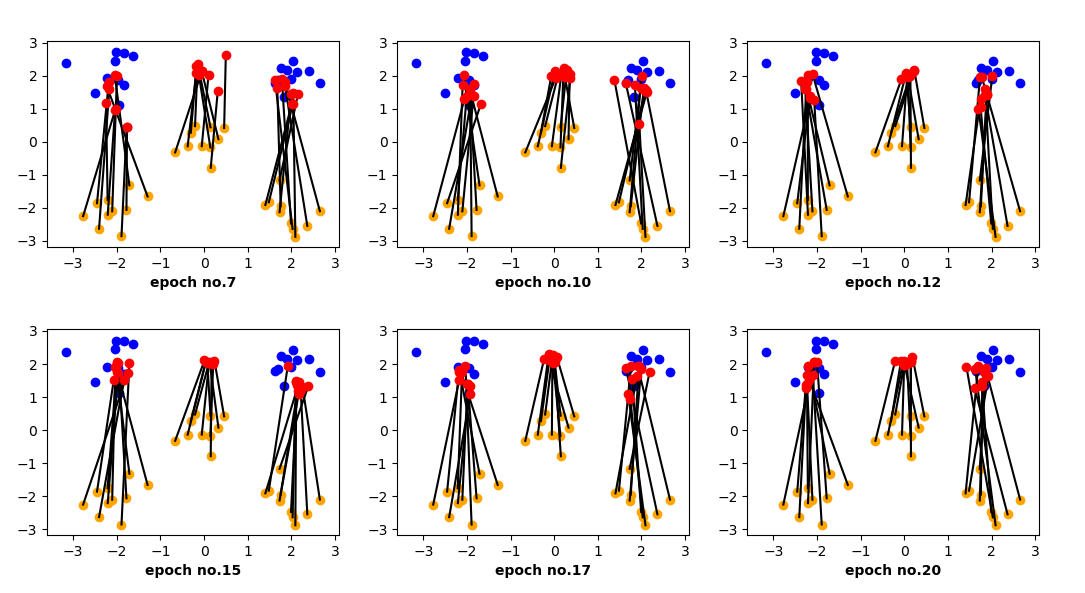}
\caption{Early stopping of the optimization after a finite number of epochs. The results are very consistent and stable even of we stop the algorithm very early.}
\label{fig:simple_synthetic_data_early_stopping}
\end{figure*}

\subsection{Early stopping of the optimization}
\begin{figure*}[!ht]
\centering
\includegraphics[width=1\textwidth]{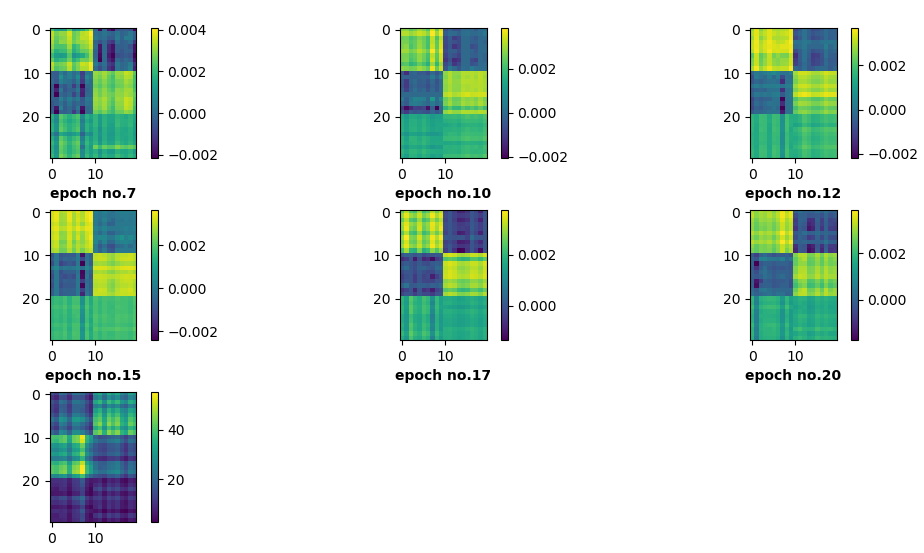}
\caption{Consistency of the transport maps with the transport costs (shown at the last row) when using different finite number of epochs. Thus, early stopping can be useful for efficiency purposes. }
\label{fig:simple_synthetic_data_early_stopping_transport_map}
\end{figure*}

We study the early stopping of our  optimization procedure. We use  the data in Fig. \ref{fig:simple_synthetic_data_diverse} and investigate the results with different number of epochs. Here, we employ the OT-SON with proper parameterization, i.e., the results shown in the forth column and the second row for OT-SON in Fig.  \ref{fig:simple_synthetic_data_diverse}. In the experiments in Fig. \ref{fig:simple_synthetic_data_diverse} we performed the optimization with $20$ epochs. Here, we study early stopping, i.e., we study the quality of results if we stop after a smaller number of epochs. According to the results in Fig. \ref{fig:simple_synthetic_data_early_stopping}, we observe that even after a small number of epochs, we obtain reliable and stable results that represent well the ultimate solution. Such a property is very important in practice, as it can significantly reduce the heavy computations. Fig. \ref{fig:simple_synthetic_data_early_stopping_transport_map} illustrates the transport maps for different number of epochs. The different transport maps at different number of epochs are consistent with the transport cost shown in the last row of Fig. \ref{fig:simple_synthetic_data_early_stopping_transport_map}. 


\subsection{Diverse classes in the source}

\begin{figure*}[!ht]
\centering
\includegraphics[width=0.333\textwidth]{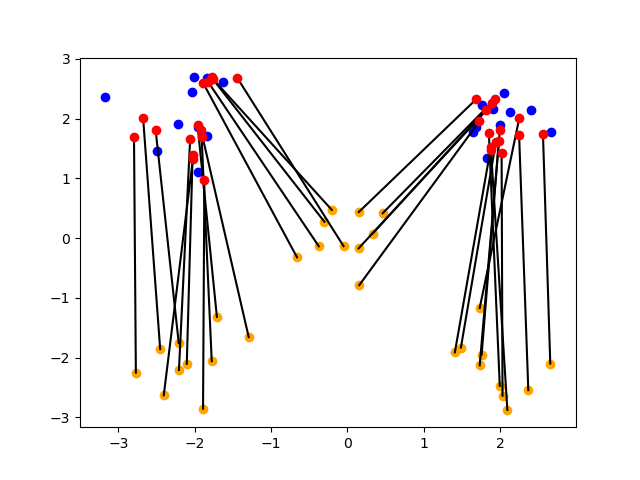}
\includegraphics[width=0.333\textwidth]{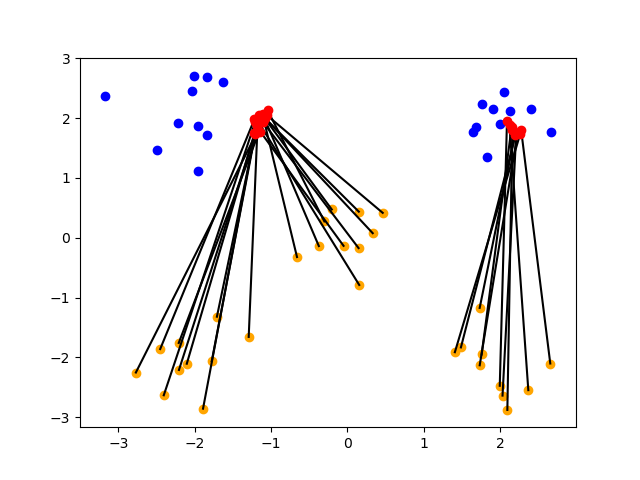}
\includegraphics[width=0.333\textwidth]{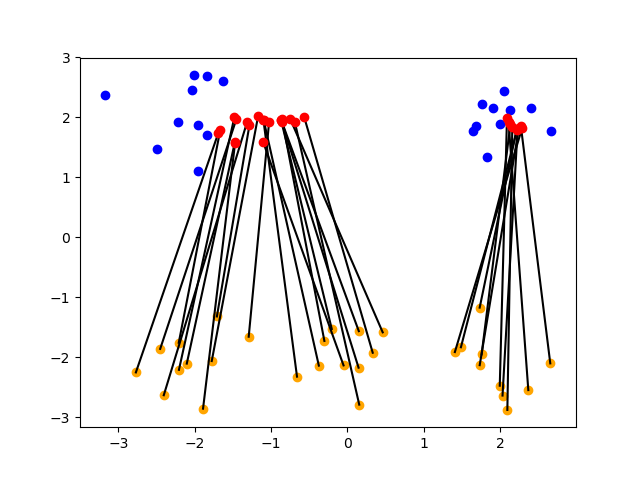}
\caption{The impact of SON regularization when the class members are diverse. The plot on the left (where $\lambda_1 = \lambda_2 =0$) performs transportation solely based on pairwise distances, thus fails to transfer the classes properly. Our SON regularization (either $\lambda_1 = 10, \lambda_2 = 0.01$ or $\lambda_1 = 100, \lambda_2 = 0.01$) improves the transportation by enforcing block-specific transfers.}
\label{fig:simple_synthetic_data_diverse_class}
\end{figure*}

We next study the case where two of the three source classes have the same label, as shown in Fig. \ref{fig:simple_synthetic_data_diverse_class}. In the source data (shown by yellow), the left and the middle data clouds have the same class labels. This example shows why the transport based on only the pairwise distances between the source and target data is insufficient. In Fig. \ref{fig:simple_synthetic_data_diverse_class}, the  left plot corresponds to $\lambda_1 = \lambda_2 =0$, the middle plot corresponds to $\lambda_1 = 10, \lambda_2 = 0.01$, and the right plot corresponds to $\lambda_1 = 100, \lambda_2 = 0.01$. We observe that the left plot (with $\lambda_1 = \lambda_2 =0$) fails to perform a proper transport of the source data. On the other hand, with incorporating our proposed regularization, the two different classes (even-though one of them is diverse) are properly transported to the target domain. We observe this kind of transfer in both of the middle ($\lambda_1 = 10, \lambda_2 = 0.01$) and right ($\lambda_1 = 100, \lambda_2 = 0.01$) plots. 

\subsection{Fewer classes in the source}
\begin{figure*}[t!]
\centering
\includegraphics[width=1\textwidth]{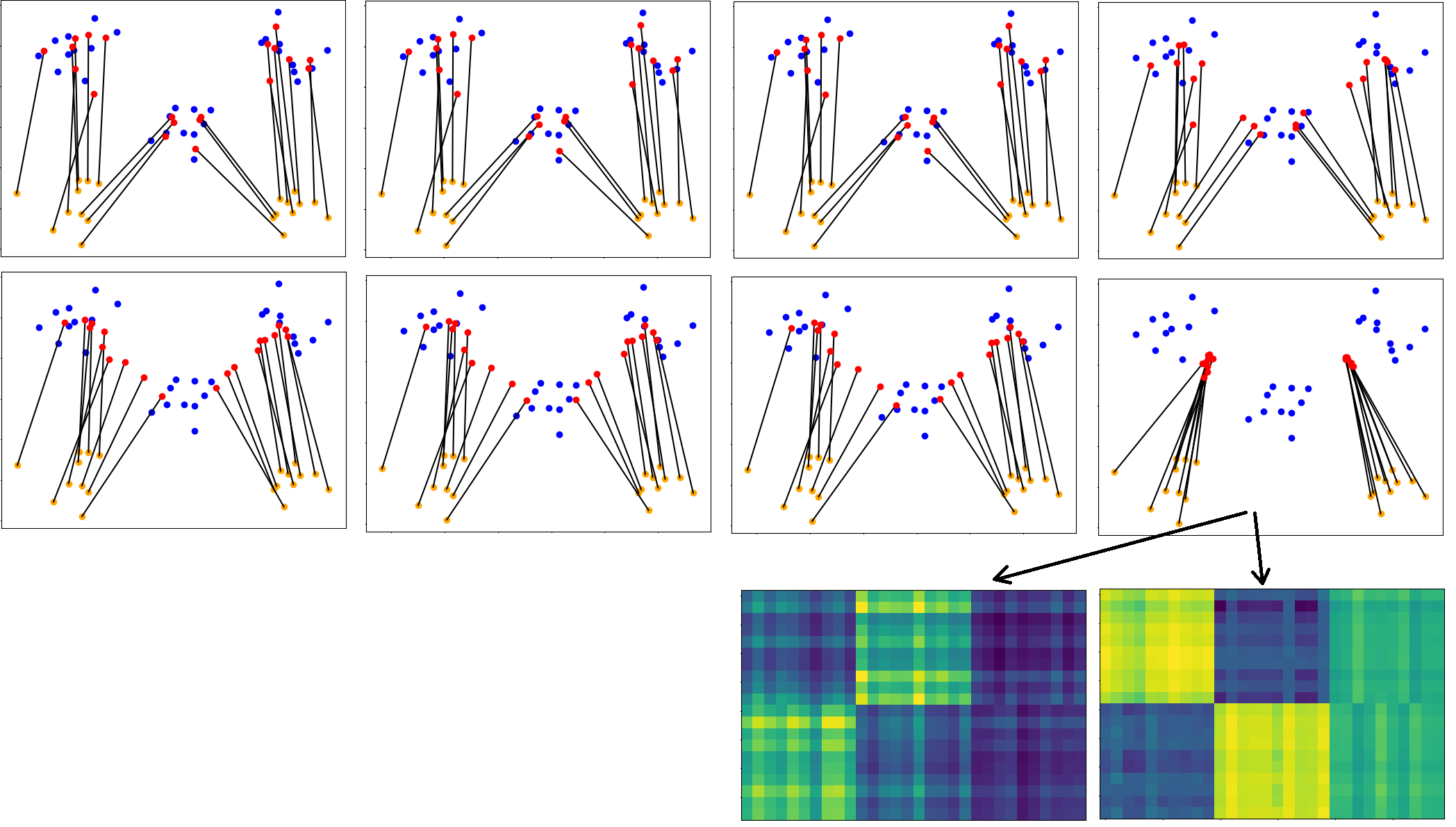}
\caption{Performance of different methods when the source has two classes and the target consists of three classes. The columns in order represent OT-l1l2, OT-lpl1, OT-Sinkhorn and OT-SON. Among different methods, only OT-SON with high regularization prevents splitting the source data among all the three target classes. The last row shows the consistency between the mapping costs and the transport map for OT-SON with high regularization.
}
\label{fig:simple_synthetic_data_diverse_2_3}
\end{figure*}

In the experiments of Fig. \ref{fig:simple_synthetic_data_diverse}, we studied the case where the number of source classes is larger the number of target classes. Here, we consider an opposite setting: we assume two classes in the source and three classes in the target, as illustrated  in Fig. \ref{fig:simple_synthetic_data_diverse_2_3}. The source, target and transported data points are respectively shown by  yellow, blue, and red.  We use the same setting and parameters as in Fig. \ref{fig:simple_synthetic_data_diverse}, i.e., the first row corresponds to low regularization and the second row to high regularization (low regularization: $\lambda_1 = 0.01, \lambda_2=0.0$, high regularization: $\lambda_1 = 10, \lambda_2 = 5$). We observe that similar to the results in Fig. \ref{fig:simple_synthetic_data_diverse}, only OT-SON with high regularization prevents splitting the source data among all the three target classes. The last row in Fig. \ref{fig:simple_synthetic_data_diverse_2_3} indicates the consistency between the mapping costs and transport map for this setting (for OT-SON with high regularization). 

%% file: unique.tex
\section{Uniqueness}
\label{sec:uniqueness}
An elementary question concerning any optimization formulation, including the Kantorovich problem and its regularization in \eqref{eq:framework}, is the uniqueness of their optimal solution, and a standard method for verifying uniqueness is to establish strong convexity of the objective function. Even though it is seen that the objective in (\ref{eq:framework}) is not strongly convex, we are nevertheless able to identify conditions, under which the solution still remains unique. For this, we develop 
an alternative approach, which is not only useful in our framework, but may also be generically used in many similar problems including a wide range of linear programming (LP) relaxation problems, and for this reason it is first presented. Our approach is based on the following definition:
\begin{definition}
We call a (global) optimal solution $\bX_0$ of a convex optimization problem
\[
\minl_{\bX\in\scrS}\scrF(\bX),
\]
where $\scrF(\ldotp)$ is a convex function and $\scrS$ is a convex set, a {\bf resistant optimal point} if adding a linear perturbation term $\langle\tlbD,\bX\rangle$ with sufficiently small coefficients in $\tlbD$ to the objective leads to an arbitrarily small perturbation of the solution $\bX_0$. In mathematical terms for any open neighborhood $\calN$ of $\bX_0$ there exists an open neighborhood $\calM$ of $\tlbD$ around $\tlbD=\bzero$ such that
\[
\forall\tlbD\in\calM,\quad\calN\cap\arg\minl_{\bX\in\scrS}\scrF(\bX)+\langle\tlbD,\bX\rangle
\neq\emptyset.
\]
\end{definition} 
Accordingly, we have the following result:
\begin{theorem}
\label{lemma:resistance}
A resistant optimal point of a convex optimization problem is its unique optimal point.
\begin{proof}
Suppose that there exists a different optimal point $\bX^\prime$. Take $\bD_0=\frac{\bX_0-\bX^\prime}{\|\bX_0-\bX^\prime\|}$, $r=\|\bX_0-\bX^\prime\|$ and $\tlbD=\epsilon\bD_0$ for arbitrary $\epsilon>0$. Further, define $\calN$ as the ball of radius $\delta=r/2$ centered at $\bX_0$. Note that for each $\bY\in \calN$ we have 
\[
\scrF(\bY)+\langle\tlbD,\bY\rangle\geq
\scrF(\bX_0)+\langle\tlbD,\bY\rangle
=\]
\[\scrF(\bX^\prime)+\langle\tlbD,\bX^\prime\rangle+\langle\tlbD,(\bY-\bX_0)+(\bX_0-\bX^\prime)\rangle.
\]
Now, note that
$
\langle\tlbD,(\bY-\bX_0)+(\bX_0-\bX^\prime)\rangle\geq -\delta\epsilon+r\epsilon>0,
$
which establishes
\[
\scrF(\bY)+\langle\tlbD,\bY\rangle>
\scrF(\bX^\prime)+\langle\tlbD,\bX^\prime\rangle.
\]
Hence,
$
\calN\cap\arg\minl_{\bX\in\scrS}\scrF(\bX)+\langle\tlbD,\bX\rangle=\emptyset
$ and
since $\epsilon=\|\tlbD\|$ is arbitrarily small, we conclude that $\bX_0$ is not a resistant optimal point. This contradicts the assumption and shows that the solution is unique.  
\end{proof} 
\end{theorem}
Theorem \ref{lemma:resistance} is a general way to establish uniqueness. In fact, we can show that the strong convexity condition is a special case of this result:
\begin{theorem}
\label{thm3:resistant}
If $\scrF$ is continuous and strongly convex, then the global minimal point of $\scrF$ over a convex set $\scrS$ is resistant.
\begin{proof}
Denote the optimal point by $\bX^*$. By strong convexity, there exists a $\gamma>0$ such that for any feasible point $\bX\in\scrS$, we have $\scrF(\bX)-\scrF(\bX^*)\geq\frac\gamma 2\|\bX-\bX^*\|_\mathrm{F}^2$. Take $\scrG=\scrF+\langle\tlbD,\bX\rangle$ and note that $\scrG(\bX)-\scrG(\bX^*)\geq\frac\gamma 2\|\bX-\bX^*\|_\mathrm{F}^2+\langle\tlbD,\bX-\bX^*\rangle\geq \frac\gamma 4\|\bX-\bX^*\|_\mathrm{F}^2-\frac 2{\gamma}\|\tlbD\|_\mathrm{F}^2$. This shows that $\scrG>\scrG(\bX^*)$ and hence does not have any global optimal point outside the closed sphere $\{\bX\mid\|\bX-\bX^*\|_\mathrm{F}\leq\frac{\sqrt 8}\gamma\|\tlbD\|_\mathrm{F}\}$. Since $\scrG$ is continuous, it also attains a minimum inside the sphere, which then becomes the global optimal point. We conclude that for any $\epsilon>0$, taking $\|\tlbD\|<\frac{\gamma\epsilon}{\sqrt{8}}$ leads to an optimal solution inside a ball of radius $\epsilon$ centered at $\bX^*$. This shows that the solution is resistant. 
\end{proof}
\end{theorem}
{\bf Uniqueness for \eqref{eq:framework}:} One special case of resistant optimal points, that will be useful in our analysis, is when there exists a neighborhood $\calM$ of $\bzero$ such that
\[
\forall\tlbD\in\calM,\quad\bX^*\in\arg\minl_{\bX\in\scrS}\scrF(\bX)+\langle\tlbD,\bX\rangle.
\]
We call such a resistant optimal point an {\bf extremal optimal point}. 
Later, we consider an analysis where we give conditions on $\bD$ to ensure that a desired solution $\bX^*$ is achieved.  Our strategy for uniqueness in this analysis is to show that under the same conditions, the desired optimal point is also extremal and hence unique, according to Theorem 1. In the case of the problem in \eqref{eq:framework}, adding the term $\langle\tlbD,\bX\rangle$ modifies the cost matrix $\bD$ to $\bD+\tlbD$. Hence, being an extremal optimal point is in this case equivalent to the solution $\bX^*$ being maintained following a perturbation of the matrix $\bD$ in a sufficiently small open neighborhood. This is easy to achieve in our planted model analysis, because the optimality of $\bX^*$ is guaranteed by a set of inequalities on $\bD$, which remain valid under small perturbations, simply by requiring the inequalities to be strict. As seen, Theorem \ref{lemma:resistance} and extremal optimality, in particular, can be powerful tools for establishing uniqueness beyond strong convexity. 

%% file: algorithm_supp.tex
\section{Remarks on the Optimization Algorithm}
\label{sec:alg_sup}
{\bf Efficient Computation}: While the objective in (\ref{eq:fullobj}) may appear complex as it involves $n^2$ terms, the associated algorithm is stochastic and incremental, thus only involving one term in (\ref{eq:fullobj}) for each iteration,  thus greatly reducing the complexity as a result. The simplification of the algorithm is also due to the proximal update detailed in Theorem~\ref{lem:proxop} (and subsequent projection) used in each iteration update of a pair of rows or columns.  We further note that an early stopping typical of stochastic schemes is likely, making a full-run to convergence unnecessary (see Section 4.4 in \cite{bcn18}), and in practice avoiding the impact of the $n^2$ terms on the performance.  When the underlying data satisfies the structure of the stochastic block model, the problem size is essentially $B^2 \ll  n^2$, as the number of required iterations is determined by an adequate sampling of all blocks. 

{\bf Just-in-Time Update}: In our problem of interest in \eqref{eq:framework}, the number of variables quadratically grows with the problem size. For such problems, incremental algorithms may become infeasible in large-scale. Note that each iteration of our algorithm includes proximal and projection operators, that update only a small group of variables. This allows us to apply the Just-in-Time approach in \cite{schmidt2017minimizing} to resolve the problem with the number of variables.
In our problem, each term $\phi_n(x)$ and constraint $S_m$ only involves a small subset $x_{I_n}:= (x_i, i \in I_n)$ of the variables, where $I_n \subseteq [D]$. Hence, the projection and proximal operators alter only a small subset of variables, dramatically reducing the amount of computation. We exploit this to give an algorithm that has much cheaper per-iteration cost. Note that the vanilla algorithm explained in \eqref{eq:alg1} and \eqref{eq:alg2} still operates on the full set of variables as the memory vectors become non-sparse by the updating rule in \eqref{eq:alg2}. We resolve this issue by following the Just-in-Time approach in \cite{schmidt2017minimizing} and modifying \eqref{eq:alg2} to
\begin{equation}\label{eq:alg2prime}
\ba_t=\rho\frac{\bx_t-\bx_{t+1}}{\mu}-\alpha\left(\suml_n\bg_n+\suml_m\bh_m\right)_{I_t}
\end{equation}
where $I_t$ denotes the set of variables involved in the $t^\tth$ iteration and we define $(\by)_I$ for a vector $\by=(y_1,y_2,\ldots,y_d)$ as a vector $\by^\prime=(y_1^\prime,y_2^\prime,\ldots,y_d^\prime)$ such that
\[
y_i^\prime=\left\{\begin{array}{cc}
    \frac{Ky_i}{K_i} & i\in I \\
     0 & i\notin I
\end{array}\right.
\]
where $K=M+N$ and $K_i$ is the number of objective terms $\phi_n$ and constraint sets $S_m$ including the $i^\tth$ variable $x_i$.